%% file: paper.tex
\documentclass{article}
\usepackage{log_2024}						

\usepackage{booktabs}						
\usepackage{multirow}						
\usepackage{amsfonts}						
\usepackage{graphicx}						

\usepackage[numbers,compress,sort]{natbib}

\usepackage{algorithm}
\usepackage[noend]{algorithmic}
\usepackage{subfig}
\usepackage{booktabs}
\usepackage[export]{adjustbox}
\usepackage{blindtext}
\usepackage{titlesec}
\usepackage{amsmath}
\usepackage{amssymb}
\usepackage{mathtools}
\usepackage{amsthm}
\newcommand{\myparagraph}[1]{\textbf{#1}\ }
\input{math_commands}

\newcommand{\LOO}{\mathrm{LOO}}
\usepackage{wrapfig}
\usepackage{todonotes}
\usepackage[toc,page]{appendix}

\newcommand{\resumetocwriting}{%
  \addtocontents{toc}{\protect\setcounter{tocdepth}{\arabic{tocdepth}}}}

\title[Ising on the Graph: Task-specific Graph Subsampling via the Ising Model]{Ising on the Graph: Task-specific Graph Subsampling via the Ising Model}

\author[Maria Bånkestad]{%
Maria Bånkestad\\
Uppsala University\\
RISE\\
\email{maria.bankestad@ri.se}\And
Jennifer R. Andersson\\ 
Uppsala University\\
\email{}\And
Sebastian Mair\\ 
Linköping University\\ Uppsala University\\
\email{}\And
Jens Sjölund\\
Uppsala University\\
\email{}
}

\begin{document}

\maketitle

\begin{abstract}
Reducing a graph while preserving its overall properties is an important problem with many applications. Typically, reduction approaches either remove edges (sparsification) or merge nodes (coarsening) in an unsupervised way with no specific downstream task in mind. In this paper, we present an approach for subsampling graph structures using an Ising model defined on either the nodes or edges and learning the external magnetic field of the Ising model using a graph neural network. Our approach is task-specific as it can learn how to reduce a graph for a specific downstream task in an end-to-end fashion without requiring a differentiable loss function for the task. We showcase the versatility of our approach on four distinct applications: image segmentation, explainability for graph classification, 3D shape sparsification, and sparse approximate matrix inverse determination.
\end{abstract}

\section{Introduction}\label{submission}
Hierarchical organization is a recurring theme in nature and human endeavors \citep{pumain2006}. Part of its appeal lies in its interpretability and computational efficiency, especially on uniformly discretized domains such as images or time series \citep{duhamel1990fast,mallat1999wavelet,hackbusch2013multi,lindeberg2013scale}. On graph-structured data, it is, however, no longer as clear what it means to, e.g., \emph{``sample every second point''}. Solid theoretical arguments favor a spectral approach \citep{shuman2015multiscale}, wherein a simplified graph is found by maximizing the spectral similarity to the original graph \citep{jin2020graph}. On the other hand, instead of retaining as much of the original information as possible, it may often be more valuable to distill the critical information for the task at hand \citep{cai2021graph,jin2022condensing}. The two main approaches for graph simplification, coarsening, and sparsification, involve removing nodes and edges. However, it is impossible to tailor coarsening or sparsification to a particular task using existing methods.
\looseness=-1

\begin{figure}[t]
    \centering
    \includegraphics[width = \linewidth]{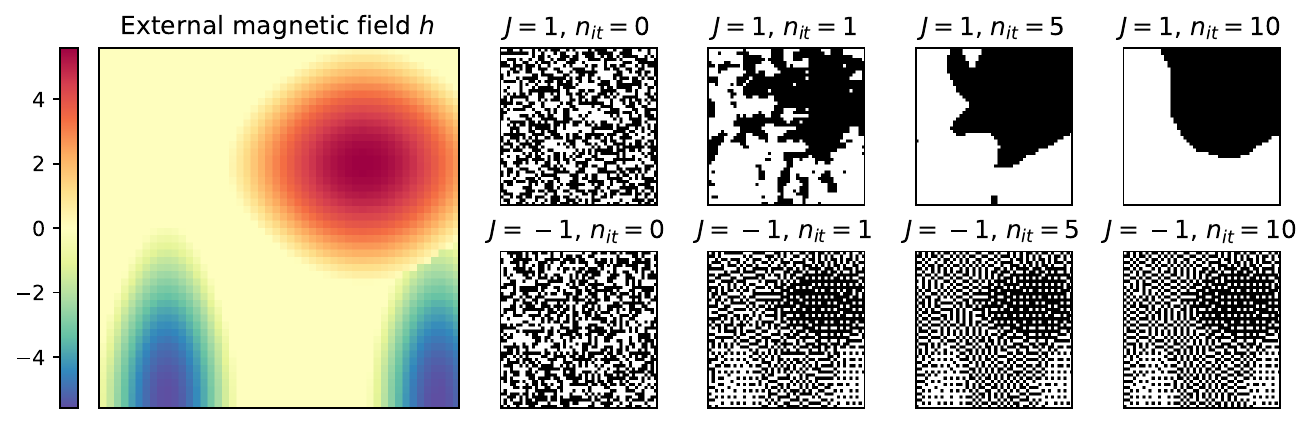}
    \caption{The external magnetic field $h$ (left) can vary spatially and influences the sampling probability relative to its strength. The sign of the coupling constant $J$ determines whether neighboring spins attract (top right) or repel each other (bottom right).}
    \label{fig:sampling_external_magnetic}
\end{figure}

We take inspiration from the well-known Ising model in physics \citep{cipra1987introduction}, which emanated as an analytical tool for studying magnetism but has since undergone many extensions \citep{wu1982potts,nishimori2001statistical}. Within the Ising model, each location is associated with a binary state (interpreted as pointing up or down), and the configuration of all states is associated with an energy. Thus, the Ising model can be seen as an energy-based model with an energy function comprising a pairwise term and a pointwise ``bias'' term. As illustrated in Figure~\ref{fig:sampling_external_magnetic}, the sign of the pairwise term determines whether neighboring states attract or repel each other, while the pointwise term (corresponding to the local magnetic field) controls the propensity of a particular alignment.
\looseness=-1

In this paper, we consider the Ising model defined on a graph's nodes or edges and augment it with a graph neural network that models the local magnetic field. There are no particular restrictions on the type of graph neural network, which means that it can, e.g., process multidimensional node and edge features. Since a node (or edge) is either included or not, we use techniques for gradient estimation in discrete models to train our model for a given task. Specifically, we show that the two-point version of the REINFORCE Leave-One-Out estimator \citep{rloo_shi} allows non-differentiable task-specific losses. We demonstrate the broad applicability of our model through examples from four domains: image segmentation, explainability for graph classification, 3D shape sparsification, and linear algebra.
\looseness=-1

\section{Background}

\myparagraph{Energy-based Models.}
Energy-based models (EBMs) are defined by an energy function $E_\theta \colon \sR^d \to \sR$, parameterized by $\theta$, where \( d \) represents the dimensionality of the input space. The density defined by an EBM is given by the Boltzmann distribution \cite{landau2013course}
\begin{equation}
    p_\theta(x) = Z_\theta^{-1} \exp(- \beta E_\theta (x)),\label{eq:boltzmann}
\end{equation}
where $\beta>0$ is the inverse temperature and the normalization constant $Z = \int \exp(-\beta E_\theta (x)) \mathrm{d}x $, also known as the partition function, is typically intractable.

A sample with low energy will have a higher probability than a sample with high energy. Many approaches have been proposed for training energy-based models \citep{song2021train}, but most of them target the generative setting where a training dataset is given and training amounts to (approximate) maximum likelihood estimation.

\myparagraph{Ising Model.}
The Ising model corresponds to a deceptively simple energy-based model that is straightforward to express on a graph $\gG = (\gV, \gE)$ consisting of a tuple of a set of nodes $\gV$ and a set of edges $\gE \subseteq \gV \times \gV$ connecting the nodes. Specifically, the Ising energy is of the form \cite{nishimori2001statistical}
\begin{equation}
    E(x) = -\sum_{(i,j)\in\gE}J_{ij}x_ix_j -  \sum_{i\in\gV}h_i x_i,\label{eq:ising_energy}
\end{equation}
where the nodes (indices $i,j$) are associated with spins $x_i\in\{\pm1\}$, the interactions $J_{ij}\in \sR$ determine whether the behavior is ferromagnetic ($J>0$) or antiferromagnetic ($J<0$), and $h_i$ is the external magnetic field.

It follows from \Eqref{eq:ising_energy} that in the absence of an external magnetic field, neighboring spins strive to be parallel in the ferromagnetic case and antiparallel in the antiferromagnetic case. An external magnetic field that can vary across nodes influences the sampling probability of the corresponding nodes. The system is disordered at high temperatures (small $\beta$) and ordered at low temperatures.

\myparagraph{Graph Neural Network.}
A graph neural network (GNN) consists of multiple message-passing layers \citep{gilmer2017neural}.
Given a node feature $x_i^k$ at node~$i$ and edge features $e_{ij}^k$ between node~$i$ and its neighbors $\{j \colon j \in \mathcal{N}(i)\}$, the message passing procedure at layer~$k$ is defined as
\looseness=-1
\begin{equation}
    m_{ij}^k = f^\text{m}(x_i^k, x_j^k, e_{ij}^k), \ \ \ 
    \hat{x}_i^{k+1} = f^\text{a}_{j \in \mathcal{N}(i)} (m_{ij}^k), \ \ \ 
    {x}_i^{k+1} = f^\text{u}(x_i^k, \hat{x}_i^{k+1}),
\end{equation}
where $f^\text{m}$ is the message function, deriving the message from node~$j$ to node~$i$, and $f^\text{a}_{j \in \mathcal{N}(i)}$ is a function that aggregates the messages from the neighbors of node $i$, denoted $\mathcal{N}(i)$. The aggregation function $f^\text{a}$ is often just a simple sum or average. Finally, $f^\text{u}$ is the update function that updates the features for each node. A GNN consists of message-passing layers stacked onto each other, where the node output from one layer is the input of the next layer.

Note that we can rewrite the energy in \Eqref{eq:ising_energy} to reveal the message-passing structure of the Ising model as
\begin{equation}\label{eq:energy_eff}
    E(x^k) = -\sum_{i\in\gV} x_i^k \cdot \left ( h_i + \sum_{j\in \mathcal{N}(i)} J_{ij} x_j^k \right )
    = -\sum_{i\in\gV} x_i^k h^\text{eff}_i(x^k), \ \ \ \ h^\text{eff}_i(x^k) = h_i + \sum_{j\in \mathcal{N}(i)} J_{ij} x_j^k,
\end{equation}
where \(h^\text{eff}_i(x^k)\) is the effective magnetic field at the \(k\)-th iteration.
\Eqref{eq:energy_eff} can be viewed as a message-passing step in a GNN, where the message function \(f^\text{m}\) is \(m_{ij}^k = J_{ij} x_j^k\), the aggregation function \(f^\text{a}\) is the sum \(\hat{x}_i^{k+1} = \sum_{j\in \mathcal{N}(i)} m_{ij}^k\), and the update function \(f^\text{u}\) is \({x}_i^{k+1} = -x_i^k \left( h_i + \hat{x}_i^{k+1} \right)\). Here, \( h_i \) is the external magnetic field at node \( i \), while \( x_i^k \) is the spin state of node \( i \) at iteration \( k \). The spin state updates based on \( h_i \) and the states of neighboring nodes \( \{x_j^k\}_{j \in \mathcal{N}(i)} \).

\myparagraph{Gradient Estimation in Discrete Models.}\label{sec:gradient_estimation}
Let $p_\theta(x)$ be a discrete probability distribution and assume that we want to estimate the gradient of a loss defined as the expectation of a loss $\ell(x)$ over this distribution, i.e., 
\begin{equation}\label{eq:grad_exp}
    \mathcal{L}(\theta) = \mathbb{E}_{x\sim p_{\theta}(x)}\left [\ell(x) \right].
\end{equation}
Because of the discreteness, it is impossible to compute a gradient directly using backpropagation. Through a simple algebraic manipulation, we can rewrite this expression as 
\begin{equation}
    \nabla_\theta\mathbb{E}_{x\sim p_\theta(x)}\left[\ell(x) \right] = \mathbb{E}_{x\sim p_\theta(x)} \left[ \ell(x) \nabla_\theta \log p_\theta (x) \right],
\end{equation}
which is called the REINFORCE estimator \citep{williams1992simple}. Though general, it suffers from high variance. A way to reduce this variance is to use the REINFORCE Leave-One-Out (RLOO) estimator \citep{rloo_shi}, 
\begin{equation}\label{rloo_eq}
    {(\nabla_\mathbf{\theta}\mathcal{L})}_{\LOO}^K = \frac{1}{K} \sum_{k=1}^K\left (\ell(x^{(k)})-\bar{\ell}_j \right)\nabla_\theta \log p_\theta (x^{(k)}), \ \text{ where } \bar{\ell}_j=\frac{1}{K-1}\sum_{j=1, j \neq k}^K\ell(x^{(j)}).
\end{equation}

\section{Method}\label{sec:method}
In GNN terminology, the interactions \( J_{ij} \) represent edge features, while the external magnetic fields~\( h_i \) serve as node features. The spin state of each node \( x_i \) is a dynamic variable that can take values in \(\{\pm1\}\). In particular, we use a GNN to parameterize the external magnetic field, $h_\theta\colon\mathcal{G}\to \mathbb{R}^{|\gV|}$, defining the Ising model according to \Eqref{eq:boltzmann}. The final output is a binary partitioning of the input graph, given by sampling this energy-based model.

\myparagraph{Sampling.}
We use the Metropolis-Hastings algorithm \cite{mackay2003information} for the sampling, see Algorithm~\ref{alg:metro_ising}. However, to make it computationally efficient, we parallelize it by first coloring the graph so that nodes of the same color are never neighbors. 
Then, we can perform simultaneous  Metropolis-Hastings updates for all nodes of the same color. A simple checkerboard pattern perfectly colors grid-structured graphs, such as images. For general graphs, we run a greedy graph coloring heuristic \citep{miller1999optimal} that produces a coloring with a limited, but not necessarily minimal, number of colors $C$ (see Appendix \ref{app:graph_coloring}). \looseness=-1

\begin{algorithm}[t]
\begin{small}
   \caption{Monte Carlo Sampling of the Ising Model}
   \label{alg:metro_ising}
   \begin{algorithmic}
   \small
   \STATE {\bfseries Input:} Graph $\{\mathbf{G}_c \}_{c=1}^C$ grouped by color $c$, interactions term $J$, external magnetic field $h$, iterations $T$.
   \STATE {\bfseries Output:} State configuration $x \in \{\pm1\}^{|\gV|}$.
   \FOR{$i=1$ {\bfseries to} $|\gV|$}
        \STATE $x_i \sim U(\{-1,1\}$) \hfill \COMMENT{Sample an initial state}
   \ENDFOR
   \FOR{$t=1$ {\bfseries to} $T$}
       \FOR{$c=1$ {\bfseries to} $C$}
           \FOR{{\bfseries all nodes} $ x_i \in \mathbf{G}_c$}
               \STATE{$\Delta E_i  = 2x_i\left( J\sum_{j\in \mathcal{N}(i)}x_j + h_i\right )$}
               \STATE{$r \sim U([0,1])$} \hfill \COMMENT{Sample $r$ uniformly at random}
               \IF{$\Delta E_i <0$ {\bfseries or} $\exp(-2\beta \Delta E) > r$}
                   \STATE{$x_i = -1 \cdot x_i$} \hfill \COMMENT{Flip the spin of node $i$}
               \ENDIF
           \ENDFOR
        \ENDFOR
   \ENDFOR
\end{algorithmic}
\end{small}
\end{algorithm}

\myparagraph{Controlling the Sampling Fraction.}\label{sec:sampling_fraction}
For some tasks, the obvious way to minimize the loss is to either sample all nodes or none. To counteract this, we need a way to control the fraction of nodes to sample or, equivalently, to control the average magnetization $\eta$, which is defined as  
\begin{align}\label{eq:eta_def}
    \eta &= 
    \frac{1}{|\mathcal{V}|}\sum_{i=1}^{|\mathcal{V}|} \mathbb{E}_{x\sim p_\theta(x)}\left[x_i\right],
\end{align}
where $x\in\{\pm1\}^{|\gV|}$ denotes the vector containing all states and $x_i$ is the $i$th entry in $x$.
Let $\bar{x}_i=\mathbb{E}_{x\sim p_\theta(x)}\left[x_i\right]$ denote the local magnetization at node~$i$.
Since the Markov blanket of a node is precisely its neighborhood, we can use \Eqref{eq:energy_eff} to rewrite the local magnetization as
\begin{equation}
    \bar{x}_i=
    \mathbb{E}_{j\in\mathcal{N}(i)}\left[\mathbb{E}_i\left[x_i \mid x_j\right]\right]
    =\mathbb{E}_{j\in\mathcal{N}(i)}\left[\tanh(\beta h^\text{eff}_i(x))\right].\label{eq:local_magnetization_exact}
\end{equation}

The corresponding mean-field (variational) approximation is a nonlinear system of equations in $\bar{x}$ \citep{mackay2003information}:
\looseness=-1
\begin{equation}\label{eq:mean-field}
    \bar{h}^\text{eff}_i(\bar{x})=h_i+\sum_{j\in\mathcal{N}(i)}J_{ij}\bar{x}_j, \ \ \ \ 
        \bar{x}_i = \tanh(\beta \bar{h}_i^\text{eff}).
\end{equation}
Solving this yields a deterministic way of approximating the average magnetization $\eta$. 
However, we are primarily interested in the ordered regime, where $\beta \bar{h}^\text{eff}_i$ is large, and it approximately holds that $x_j=J_{ij}x_i$, which implies that $\bar{x}_i=\tanh(\beta h_i)$ and, thus,
\begin{equation}
        \Tilde{\eta}_\text{det}=\frac{1}{|\mathcal{V}|}\sum_{i=1}^{|\mathcal{V}|}\tanh\left(\beta h_i\right).\label{eq:reg_deterministic}
\end{equation}
This can be contrasted with the stochastic estimate we get from using a one-sample Monte Carlo estimate in \Eqref{eq:local_magnetization_exact}:
\looseness=-1
\begin{equation}
    \Tilde{\eta}_\text{sto}=\frac{1}{|\mathcal{V}|}\sum_{i=1}^{|\mathcal{V}|}\tanh\left(\beta h^\text{eff}_i(x))\right),\quad x\sim p_\theta(x).\label{eq:reg_stochastic}
\end{equation}

\myparagraph{Learning.}
In contrast to most other energy-based models \citep{song2021train}, ours is not an unconditional generative model. Neither is it a supervised regression task \citep{gustafsson2020train}, since we \emph{learn to subsample} without requiring ground truth labels on which nodes are the correct ones to choose. Instead, we target the scenario where it is challenging to specify useful subgraphs for a downstream task but we have access to an associated, possibly non-differentiable, loss (which may use labels for the downstream task, as in Section~\ref{sec:explain}). However, supervised training is also possible, as we demonstrate in Section~\ref{sec:segmentation}.

We need the gradient of the log probability for gradient-based training of the GNN in our Ising model, which can be decomposed as a sum of two terms:
\begin{equation}
    \nabla_\theta \log p_\theta(x) = -\beta \nabla_\theta E_\theta (x) - \nabla_\theta \log Z_\theta.\label{eq:grad_log_prob}
\end{equation}
We are particularly interested in training the model for a downstream task associated with a loss $\ell(x)$ defined in terms of samples $x\sim p_\theta(x)$ from the Ising model. The goal is then to minimize the expected loss as defined in \Eqref{eq:grad_exp}.

Since we are operating in a discrete setting, we use the gradient estimation technique described in Section~\ref{sec:gradient_estimation}. Specifically, using $K=2$ in \Eqref{rloo_eq}, the problematic $\log Z_\theta$ terms cancel in the RLOO estimator, and we obtain
\begin{equation}
    {(\nabla_\mathbf{\theta}\mathcal{L})}_{\LOO}^2
   =-\frac{\beta}{2}\left(\ell(x^{(1)})-\ell(x^{(2)})\right) 
    \cdot \nabla_\theta  \left(E_{\theta}(x^{(1)})-E_{\theta}(x^{(2)}) \right ).\label{eq:ising_rloo}
\end{equation}
To train the model, we can thus draw two independent samples $x^{(1)}, x^{(2)} \sim p_\theta (x)$ from the Ising model and estimate the gradient using \Eqref{eq:ising_rloo}, where $\nabla_\theta  \left(E_{\theta}(x^{(1)})-E_{\theta}(x^{(2)}) \right)$ can be computed by automatic differentiation.

To control the sampling fraction, as described in Section~\ref{sec:sampling_fraction}, we could either include a regularization term that depends on the stochastic sampling fraction (\Eqref{eq:reg_stochastic}) directly in the task-specific loss~$\ell$, or use a deterministic approximation based on \Eqref{eq:reg_deterministic} to move it outside the expectation and thereby enable direct automatic differentiation. Empirically, we found the latter option to work better. Specifically, we use the training loss 
\begin{align} \label{eq:loss_with_reg}
    \mathcal{L}_{\text{tot}}(\theta)   
    &=\mathbb{E}_{x\sim p_{\theta}(x)}\left [\ell(x) \right ] +  (\Tilde{\eta}_\text{det}(\theta)-\eta)^2,
\end{align}
where the desired value of the average magnetization $\eta$ is defined as in \Eqref{eq:eta_def}.

\begin{figure}[t]
    \centering
    \includegraphics[width = 0.87\linewidth]{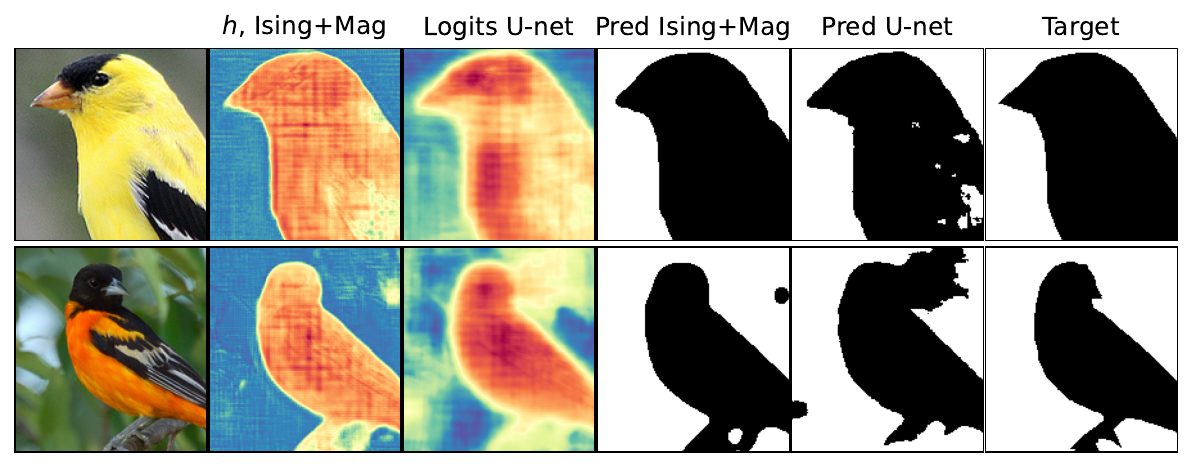}
    \caption{Comparison of a U-Net trained with a cross-entropy loss against an Ising model with a learned magnetic field. From left to right are the image, the Ising model magnetic field, the U-Net output (the logits), the Ising model prediction, the U-Net prediction, and the true segmentation mask.}
    \label{fig:segmentation}
\end{figure}

\newpage

\section{Applications}\label{sec:application}

We apply the proposed energy-based graph subsampling method across four distinct areas: image segmentation, explainability for graph classification, 3D shape sparsification, and sparse approximate matrix inverses. Code is available at \url{https://github.com/mariabankestad/IsingOnGraphs}.

\subsection{Illustrative Example on Image Segmentation}\label{sec:segmentation}
To provide an intuitive visual demonstration, we first apply our approach to an image segmentation task, even though our main objective is learning Ising models on general graphs. An image can be represented as a graph, where pixels are nodes and edges denote neighboring pixel relationships \citep{geman1984stochastic}. In the ferromagnetic case (\( J = 1 \)), the Ising model encourages nearby pixels to have similar values, resulting in smooth, coherent segmentations. Using a probabilistic graphical model to post-process neural network outputs can significantly improve object boundary localization \citep{chen2017deeplab}. Unlike the general setting in Section~\ref{sec:method}, this example has ground-truth labels (segmentation masks), allowing for a supervised approach with no downstream task involved.

\myparagraph{Our Approach.}
Because of the grid structure, the message from the 8-connected neighbors can be implemented as a convolutional layer with kernel size three and zero padding. Specifically, we define the energy per pixel as:
\[
    E_i = - x_i \left (  \text{Conv}_W \left ( x_i\right) + h_\theta (x_i) \right ), \ \ \ \ W = \begin{bmatrix}
        w^*&w& w^*\\
        w^{\phantom{*}}&0&w^{\phantom{*}}\\
        w^*&w& w^*       
    \end{bmatrix}.
\]
Here, $ \text{Conv}_W$ is an image convolutional operator with weights $w$, and $h_\theta (x_i)$ is an external magnetic field with learnable parameters $\theta$. The weights $w$ and $w^*$ have the same sign but can have different values. In contrast to \citet{zheng2015conditional}, who use the mean-field approximation in \Eqref{eq:mean-field} to learn a fully-connected conditional random field end-to-end, our approach applies to graph-structured energy functions and allows task-specific losses.

Since we consider binary segmentation, we use a pixel-wise \emph{supervised} misclassification loss
\begin{equation}
    \ell(x) = \frac{1}{|\gV|}\sum_{i=1}^{|\gV|} |x_i-y_i|, \ \text{ where the target value is }\ y_i \in \{-1,1\}.
\end{equation}

\myparagraph{Results.}
We use the Caltech-UCSD Birds-200-2011 dataset \citep{WahCUB_200_2011} and model the magnetic field of the Ising model by a U-Net \citep{ronneberger2015u}. See Appendix~\ref{app:unet} for details and additional results. Directly training the model for segmentation using a standard cross-entropy loss yields an average Dice score of 0.857 on the test set. In contrast, using a learned magnetic field in the Ising model, the average Dice score increases to 0.870 on the test set. Figure \ref{fig:segmentation} compares two predictions from the two models. Notably, the output from the magnetization network of the Ising model exhibits clearer and sharper borders compared to the logits predicted by the plain segmentation model. This distinction is also reflected in the predictions, where the Ising model's segmentation mask appears more even and self-consistent.

\subsection{Explainability for Graph Classification}\label{sec:explain}
GNNs integrate information from features with the graph topology. While being a key ingredient of their success, this makes their predictions challenging to explain \cite{Kakkad2023}. Nevertheless, there is a clear need for such explanations in high-stakes applications like drug design. We focus on the most studied class of problems in GNN explainability: graph classification. However, our approach, the Ising Graph Explainer (IGExplainer), is equally applicable to explain node and edge classifications.

\myparagraph{Our Approach.}
Like PGExplainer \cite{luo2020parameterized}, we derive subgraph explanations from a trained probabilistic classifier.
More specifically, we consider a subgraph to be an explanation when the model's prediction on that subgraph agrees with the prediction on the whole graph.
In the taxonomy of \citet{Kakkad2023}, this makes it a perturbation-based post-hoc method. To find such subgraph explanations, we train the external magnetic field of a ferromagnetic Ising model ($J>0$) by minimizing the cross-entropy between predictions $y\sim p(y\mid \gG)$ for the whole graph $\gG$ and predictions $y_\text{s}\sim p(y\mid \gG_\text{s})$ for sampled subgraphs $\gG_\text{s}$.
Specifically, subgraphs are sampled by subsampling nodes according to the Ising model and including all edges between those nodes.  
In the special case of binary classification, the cross-entropy reduces to the familiar expression
\begin{equation}
    H(y, y_\text{s})=- \left(p \log p_\text{s} + (1-p)\log(1-p_\text{s})\right),\label{eq:crossent}
\end{equation}
where we use the shorthand notation $p\vcentcolon= p(y=1\mid \gG)$ and $p_\text{s}\vcentcolon= p(y=1\mid \gG_\text{s})$.
To avoid the trivial solution $\gG_\text{s}=\gG$, we normalize the external magnetic field to be zero-sum.

\begin{figure}[t]
    \centering
    \includegraphics[width = 0.94\linewidth]{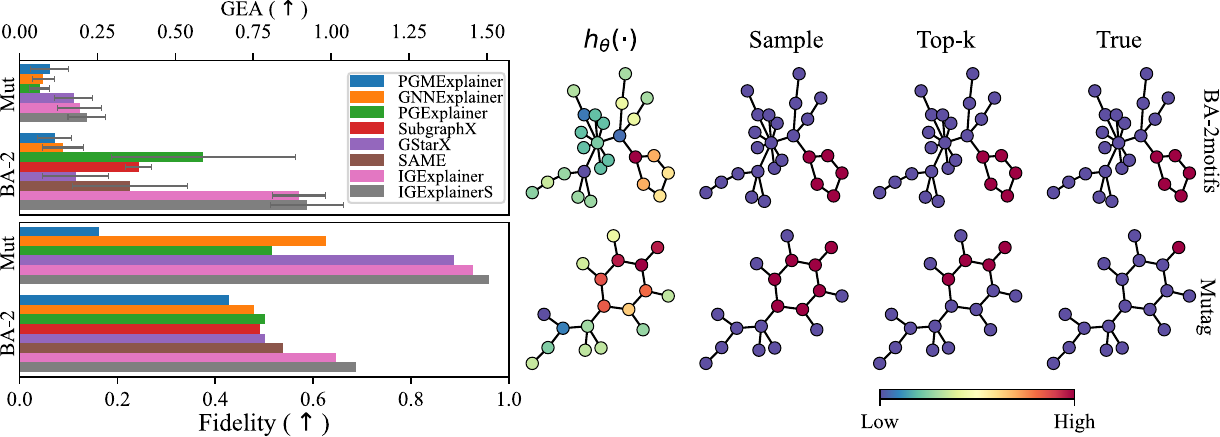}
    \caption{
    \textbf{Left:} Quantitative evaluation of IGExplainer and IGExplainerS using a synthetic dataset BA-2motifs and a real-world dataset Mutag. We report the graph explanation accuracy (GEA) and the fidelity score; higher is better for both scores. The theoretical limit of GEA is 0.43 for Mutag and 1 for BA-2motifs. \textbf{Right:} Qualitative analysis of the explanations generated by IGExplainer with the magnetic field $h$, the sample average, the top-$k$ nodes, and the truth. See Appendix \ref{app:baslines_exp} for baselines.}
    \label{fig:explain}
\end{figure}

\myparagraph{Results.} We train our IGExplainer using a three-layer graph isomorphism network (GIN) \cite{xu2018powerful} as the magnetic field $h_\theta(\cdot)$, with the cross-entropy from \Eqref{eq:crossent} as the task-specific loss. We consider two standard explainability datasets: BA-2Motifs \cite{luo2020parameterized} and Mutag \cite{kazius2005derivation,agarwal2023evaluating}. On each dataset, we first train a classification model $\mathcal{M}$ (also a GIN network), then use $\mathcal{M}$ to compute the loss for the IGExplainer via the probability distributions $p$ and $p_s$. We propose two inference procedures: IGExplainer, which uses the top-$k$ nodes according to the trained magnetic field score, and IGExplainerS, which generates an expandability mask by performing five MCMC iterations on the Ising model.
\looseness=-1

We evaluate the explanations using two metrics: graph explanation accuracy (GEA) \cite{agarwal2023evaluating} and characteristic fidelity \cite{amara2022graphframex}, which measures how well \(y_s\) aligns with \(y\). In BA-2Motifs, \(k\) is set to 5, the exact number of true nodes, while in Mutag, \(k\) is set to 40\% of the total graph size. Figure~\ref{fig:explain}~(left) illustrates that IGExplainer and IGExplainerS both outperform all baselines in terms of GEA and fidelity, with IGExplainerS further improving accuracy through the use of Monte Carlo sampling compared to IGExplainer's static top-\(k\) selection. Figure~\ref{fig:explain}~(right) visualizes the magnetic field \(h_\theta(\cdot)\) for a representative graph from each dataset, where the magnetic field strength indicates node importance. We also present the average sample from IGExplainerS, highlighting its preference for sampling cohesive node groups, contributing to higher fidelity. Appendix~\ref{app:ex_time} provides details on the inference times, showing that IGExplainerS, with an average inference time of 0.02 seconds, is approximately 1000 times faster than GStarX. Additional experimental details, model architecture, and dataset descriptions are provided in Appendix~\ref{app:graph_classsification_explainability}.

\subsection{3D Shape Sparsification}\label{sec:mesh_sparse}
An object's 3D mesh tends to have a densely sampled surface, resulting in numerous redundant vertices. These redundant vertices increase the computational workload in subsequent processing stages. Consequently, mesh sparsification (representing the object with fewer vertices and edges) is a common preprocessing step, e.g., in computer graphics \citep{wright2010sparse} and fluid dynamic simulations \citep{blazek2015computational}. The main techniques for sparsifying a mesh are based on preserving the spectral properties of the original mesh \citep{chen2022graph, keros2022generalized}. However, these methods cannot be adapted to specific downstream tasks. 

We aim to create a nested mesh using a subset of nodes from the fine mesh. Unlike methods such as QSlim \cite{garland1997surface}, which repositions vertices, our approach ensures continuity between mesh levels, a valuable feature for scientific applications (e.g., finite element analysis \cite{blazek2015computational}). Fixing vertex positions also allows for clearer comparisons of subsampling methods by isolating the effects of sampling alone.
\looseness=-1

A triangular mesh $\mathcal{M}$ is composed of vertices $\mathcal{V}$, edges $\mathcal{E}$ and faces $\mathcal{F}$, where the faces define the triangles formed by the vertices. To sparsify the mesh, we remove selected vertices and collapse the connected edges to their nearest neighboring vertex.
The distance between the vertices $\mathcal{V}^{1}$ of the original mesh and the surface of the coarser mesh $\mathcal{M}^{2}$ is defined as:
\begin{equation}\label{eq:point_mesh_dist}
    d(\mathcal{V}^{1},\mathcal{M}^{2}) = \sum_{x \in \mathcal{V}^{1}} \min_{y \in \mathcal{M}^2} ||x - y||^2_2.
\end{equation}
For each vertex in $\mathcal{M}^1$, the algorithm finds the nearest point in $\mathcal{M}^2$ and sums the squared distances.

\myparagraph{Our Approach.}
We approach mesh sparsification as a sampling problem, using the antiferromagnetic Ising model to select vertices for the coarse mesh. This model is well-suited because its ``every other'' pattern helps to distribute the samples more evenly across the mesh. Additionally, the magnetic field $h_\theta(x)$ guides sampling toward regions that are most relevant to the task, i.e., shape preservation.

\begin{figure}[t]
    \centering
    \includegraphics[width = \linewidth]{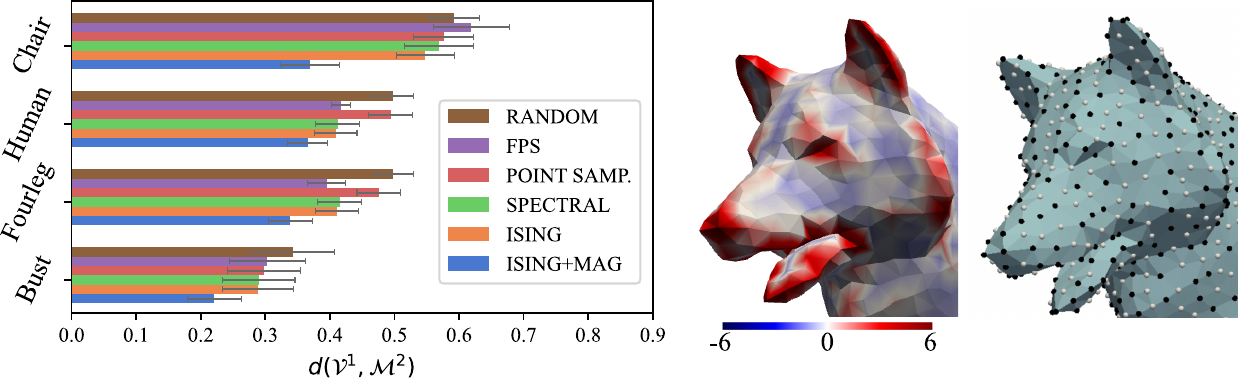}
    \caption{\textbf{Left:} Mean and standard deviation of the test vertex-to-mesh distance from the five-fold cross-validation of the four datasets. Here, lower is better. For an explanation of the other methods and the results as a table, see Appendix~\ref{app:mesh_sparse}. \textbf{Middle:} Learned magnetic field of an object. \textbf{Right:} Sampled vertices using the Ising model with a learned magnetic field. Black vertices are retained in the coarser mesh and are more important for the overall shape preservation (see the ears and nose). }
    \label{fig:dog_points}
\end{figure}

The 3D shapes are represented as graphs, with vertex positions as node features and relative distances as edge features, \( e_{ij} = x_i - x_j \). To model the magnetic field \( h_\theta(\cdot) \), we use a three-layer Euclidean GNN \citep{geiger2022e3nn}, capturing 3D geometry while remaining equivariant to rotations and translations. We also use the mesh Laplacian \citep{reuter2009discrete} to assess surface geometry (see Appendix~\ref{app:mesh_sparse}).

\myparagraph{Results.}
We use the vertex-to-mesh distance in \Eqref{eq:point_mesh_dist} as our task-specific loss and train the Ising model with $\eta=0$ in \Eqref{eq:loss_with_reg}, aiming to reduce the number of vertices in each mesh by half. As datasets, we use \emph{bust}, \emph{four-leg}, \emph{human}, and \emph{chair} \cite{chen2009benchmark}, where each dataset contains 20 different meshes. We evaluate the model using five-fold cross-validation. Figure~\ref{fig:dog_points} demonstrates that the Ising model with the magnetic field (ISING+MAG) achieves the lowest point-to-mesh distance since it is trained for this specific task. ISING and Spectral coarsening (SPECTRAL) \cite{shuman2015multiscale} perform similarly, both producing an every-other pattern, where SPECTRAL selects nodes via the graph Laplacian's largest eigenvector. While farthest point sampling \cite{eldar1997farthest,qi2017pointnet} distributes points evenly, it misses key areas. ISING+MAG combines geometric insights with effective mesh preservation. Additional details on the model, dataset, baselines, and time complexity are presented in Appendix~\ref{app:mesh_sparse}.

\begin{minipage}{\textwidth}
  \begin{minipage}{0.60\textwidth}
    \centering
    \captionsetup{type=figure}
    \includegraphics[width=0.7\textwidth]{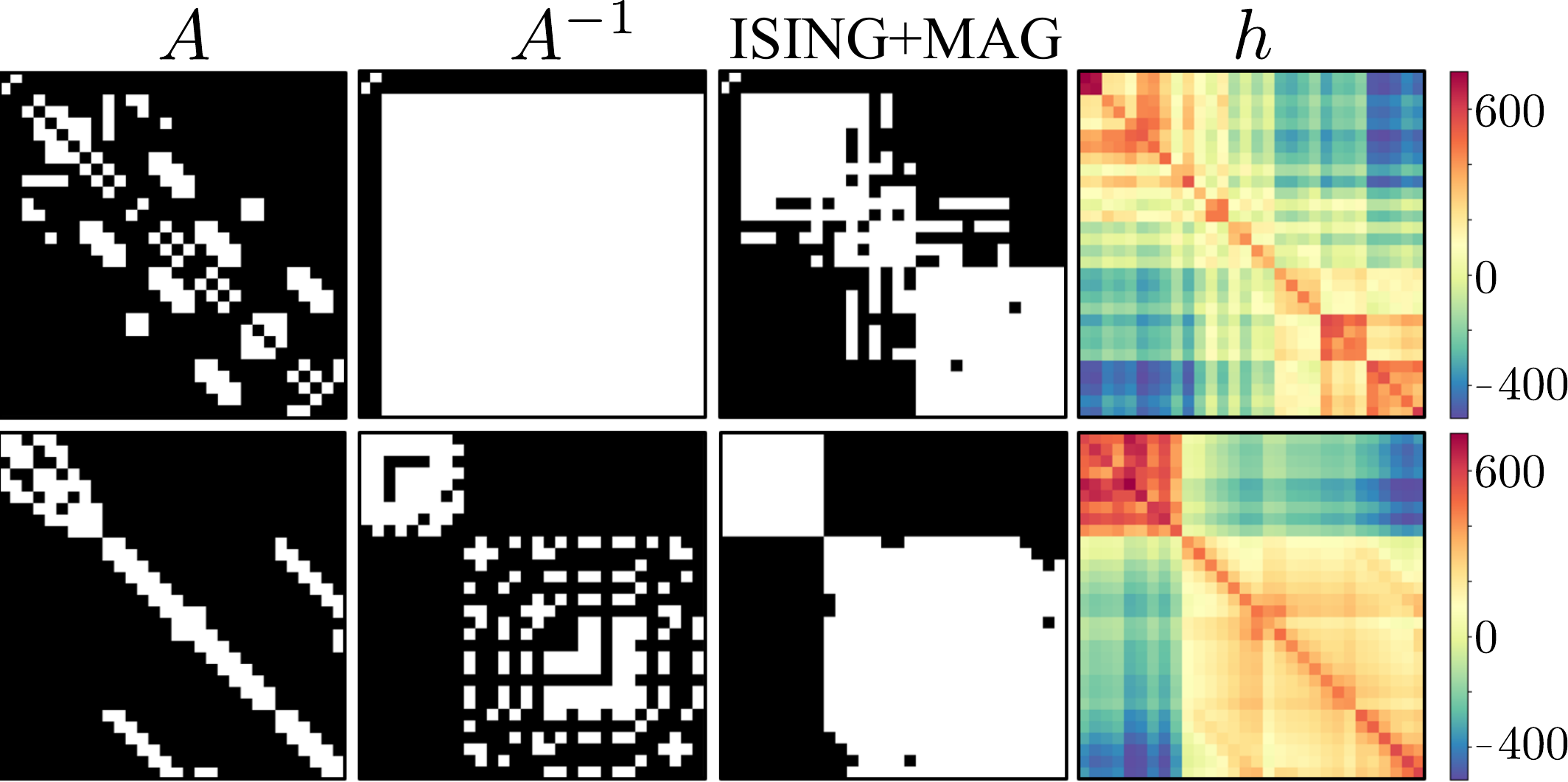}
    \captionof{figure}{Sparsity patterns of $A$, its inverse $A^{-1}$, our sparse approximate inverse (50\% elements, \emph{Setting~3}), and predicted magnetic field $h$ for two test matrices (non-zeros in white).}
    \label{fig:matrix_226_58}
  \end{minipage}
  \hfill
  \begin{minipage}{0.39\textwidth}
    \setlength{\tabcolsep}{3.5pt}
    \centering
        \captionsetup{type=table}
    \captionof{table}{Mean and standard deviation of the loss in \Eqref{eq:norm_min_loss} over the test set for each setting and model using a single sample from the learned Ising model. Lower is better.}
    \label{Table_spaic}
    \resizebox{\textwidth}{!}{
    \begin{tabular}{llll}
    \toprule
    Model &  Setting 1 &  Setting 2 &  Setting 3\\
    \midrule
    ISING+MAG & \textbf{3.28} (0.12) & \textbf{2.06} (0.26) & \textbf{1.15} (1.24)  \\
    ISING  & 3.86 (0.12) & 3.69 (0.23) & 3.78 (0.42) \\
    RANDOM & 4.04 (0.16) & 3.86 (0.18) & 3.86 (0.30) \\
    ONLY A & 4.12 (0.08) & 4.12 (0.08) & 2.00 (1.89) \\
    \bottomrule
    \end{tabular}
    }
  \end{minipage}
\end{minipage}

\subsection{Sparse Approximate Matrix Inverses}
Large linear systems $Az=b$, where $A\in \sR^{n \times n}$ is a sparse matrix, are frequently solved in many scientific and technical domains. With increasing problem size, direct methods are replaced by iterative methods, e.g., Krylov subspace methods. However, if the problem is ill-conditioned, these methods require preconditioning \cite{hausner2023neural} for which sparse approximate inverses (SAIs) are often suitable.

Our goal is to find a sparse approximate inverse $M \approx A^{-1} \in \mathbb{R}^{n \times n} $, acting as a right preconditioner for the sparse linear system $Az = b $. Let $ I $ be the identity matrix and $s \in \mathcal{S} $ denote the sparsity pattern of $ M $, with $ \mathcal{S}$ representing the space of all possible sparsity patterns. The pattern $s $ specifies the $ m $ indices of non-zero entries, while $s' \in \mathbb{R}^m $ provides their corresponding values. Further, $M'(s)$ denotes the approximate inverse with sparsity pattern $s$ and non-zeros $s'$. We find $M$ by solving
\begin{equation}\label{eq:norm_min} 
\begin{aligned}
\min_{s' \in \sR^m} \ \lVert AM'(s) - I \rVert_\mathrm{F},
\end{aligned}
\end{equation}
where $\|\cdot\|_\mathrm{F}$ denotes the Frobenius norm. However, this approach requires an a priori assumption on $s$, which affects the quality of the preconditioner. Various modifications to the described sparse approximate inverse approach exist, aiming to refine the sparsity pattern iteratively \citep{spai}.

\myparagraph{Our Approach.}
Symmetric sparse linear systems can be represented as graphs \citep{moore2023graph} where $A$ is viewed as the adjacency matrix of a weighted undirected graph where the matrix elements correspond to edge features.
The idea is to use the Ising model to sample sparsity patterns of the SAI.
However, our Ising model is defined over nodes; hence, we move the graph representation to its line graph \citep{sjolund2022graph}.
A sample $x$ from our model then corresponds to a proposed sparsity pattern $s$ of the predicted SAI~$M$.
As before, we parametrize our model by $\theta$ and learn it by minimizing \Eqref{eq:loss_with_reg} using $\eta=0$ and
\begin{equation}\label{eq:norm_min_loss} 
\begin{aligned}
l(x) = {\lVert AM(s_\theta(x)) - I \rVert}_\mathrm{F}.
\end{aligned}
\end{equation}
We aim to reduce the number of elements of an a priori sparsity pattern (possibly including all elements) by 50\%.
We obtain the specific SAI $M(s_\theta(x))$ by solving \Eqref{eq:norm_min} using the sampled sparsity pattern of the Ising model given its current parameters.
The external magnetic field is modeled by a graph GCN \citep{chen2020simple}, and we use the RLOO gradient estimator according to \Eqref{eq:ising_rloo}.
We let $A$ and $A^2$ enter as edge features.
Contrary to conventional approaches, our method is flexible in terms of a priori sparsity patterns.
Appendices \ref{app:matrix_results} and \ref{app:gcnnet} contain further details.

\myparagraph{Datasets.}
\emph{Dataset~1} consists of 1600 synthetic binary sparse matrices of size $30\times30$ and \emph{Dataset~2} consists of 1800 sparse $30\times30$ submatrices constructed from the SuiteSparse Matrix Collection \cite{matrixcollectionref}, thus resembling real-world sparse matrices.
See Appendix~\ref{app:datasets} for details.

\myparagraph{Results.}
We consider three experimental settings. \emph{Setting~1} uses $A$ and $A^2$ as edge features on \emph{Dataset~1}. Here, the a priori sparsity pattern of the SAI is the same as that of $A^2$, and our model selects 50\% of those positions for $M$. \emph{Setting~2} is similar but allows for using 50\% of all possible positions. \emph{Setting~3} follows the same approach as \emph{Setting~2} but is applied to \emph{Dataset~2}. See Appendix~\ref{app:matrix_results} for implementation details. Figure~\ref{fig:matrix_226_58} shows two test samples from Setting~3, illustrating how the learned magnetic field adapts to generate an appropriate sparsity pattern even when the true inverses have different structures. More detailed results can be found in Appendix~\ref{app:matrix_results}. Table~\ref{Table_spaic} compares the performance of our model (ISING+MAG) with three baselines: (i) an Ising model with a small constant magnetic field, tuned for the same sampling fraction as the test set (ISING), (ii) uniform random sampling from the allowed sparsity pattern with the same sampling fraction (RANDOM), and (iii) the quality of the sparse approximate inverse using only the input matrix \( A \) (ONLY~A). In all cases, ISING+MAG significantly outperforms the baselines.

\section{Related Work}

\myparagraph{Statistical Physics and Random Fields.}
The Ising model is a foundational model for magnetic systems, with generalizations such as the Potts model for multiple states \citep{wu1982potts}. The Ising model with random interactions represents spin glass systems \citep{nishimori2001statistical}, closely linked to Hopfield networks---models of associative memory that have seen renewed interest \citep{krotov2016dense,demircigil2017model,ramsauer2020hopfield}. These have, in turn, inspired the development of Markov random fields \citep{koller2009probabilistic}. Since we output the external magnetic field as an intermediate variable that defines an Ising model, it belongs to the class of conditional random fields \citep{lafferty2001conditional}, which are widely used in structured prediction tasks \citep{lample2016neural,zheng2015conditional,chen2017deeplab}.

\myparagraph{Combinatorial Optimization.}
Recent research has employed Ising models to tackle combinatorial optimization across domains such as collaborative filtering, traffic prediction, and graph learning \cite{liu2023ising, pan2023ising, wuextending}. These approaches infer coupling parameters and biases from historical data, which remain fixed post-inference. This method works well for static graphs, where inference minimizes energy configurations based on node values. In contrast, \citet{salehinejad2021pruning} apply an Ising model to enforce sparsity and prune convolutional neural networks, treating the network as a graph and using the model as a regularizer akin to dropout. Our approach differs by introducing a parametric mapping from graph data to the magnetic field, allowing for task-specific learning of Ising parameters.

\myparagraph{Graph Sparsification and Coarsening.}
Graph sparsification and coarsening aim to create a smaller graph while retaining the global structure of the original. \citet{bravo2019unifying} presents a unifying framework for both operations by reducing the graph while preserving the Laplacian pseudoinverse. Unlike our approach, theirs is algorithmic rather than learned, focusing on structural preservation without training data to adapt the reduction. Many algorithms for graph coarsening aim to preserve spectral properties \cite{safro2015advanced,loukas2018spectrally,loukas2019graph,jin2020graph}. Building on top of this idea are approaches based on neural networks \cite{cai2021graph}. A graph coarsening can also be computed based on optimal transport \cite{ma2021unsupervised}, graph fusion \cite{deng2020graphzoom}, and Schur complements to obtain embeddings of relevant nodes \cite{fahrbach2020faster}. The coarsening of graphs is also done for the scalability of GNNs \cite{huang2021scaling} and dataset condensation \cite{jin2022graph,jin2022condensing}. The goal is that GNNs perform similarly on the condensed data but are faster to train. Finally, \citet{chen2022graph} provides an overview of successful coarsening techniques for scientific computing. As for graph sparsification, spectral sparsification \cite{spielman2011spectral} and information-theoretic formulations \cite{yu2022principle} can be used. Graph partitioning can be done using quantum annealing by minimizing an Ising objective \cite{ushijima2017graph}. \citet{batson2013spectral} provides an overview on the topic of sparsification, and \citet{hashemi2024comprehensive} presents a general survey on graph reduction, including sparsification, coarsening, and condensation.

\section{Discussion and Conclusion}

We proposed a new method for graph subsampling by first learning the external magnetic field in the Ising model and then sampling from the resulting distribution. Our approach has shown potential in various applications and does not require the differentiability of the loss function of a given application. We achieved this using the REINFORCE Leave-One-Out gradient estimator and a carefully chosen regularization structure that allowed us to adjust the sampling fraction. Within a diverse set of experiments on four applications, we demonstrated the effectiveness of our method. In image segmentation, our approach generated segmentation masks with sharp borders. Our IGExplainer produced explanations with high accuracy and fidelity for graph classification. In mesh coarsening, the learned magnetic field increased the sampling rate in high-curvature regions while maintaining the characteristic ``every other'' pattern. Finally, we illustrated that our approach can learn sparsity patterns for determining sparse approximate matrix inverses via Frobenius norm minimization.

\myparagraph{Limitations and Future Work.} For our applications, the computations were fast (see Figures~\ref{fig:sim_time_only} and \ref{fig:sim_timey} in Appendix~\ref{app:mesh_sparse}), but this may change for other large-scale applications, where the sampling (incl. graph coloring) can become a bottleneck. This is something we want to address in future work. Another limitation is that the Ising model only allows for binary states, whereas some applications, e.g., image classification, require multiple classes. We believe it is possible to extend our approach to such cases by instead using the Potts model or the even more general random cluster model \citep{fortuin1972random}.

\section*{Acknowledgements}
This work was supported by the Wallenberg AI, Autonomous Systems and Software Program (WASP) funded by the Knut and Alice Wallenberg Foundation.

\bibliographystyle{unsrtnat}
\bibliography{reference}
\clearpage
\appendix
\section*{Appendix}
The appendix first presents some details on Ising sampling and graph coloring. It then continues with extra information on our four applications. We only used a single GPU, an NVIDIA GeForce RTX 3090, for all applications and performed all experiments using less than 10 GPU hours.
\resumetocwriting

\tableofcontents

\section{Ising Sampling}
\begin{figure}[ht]
    \centering
    \includegraphics[width = 0.9\linewidth]{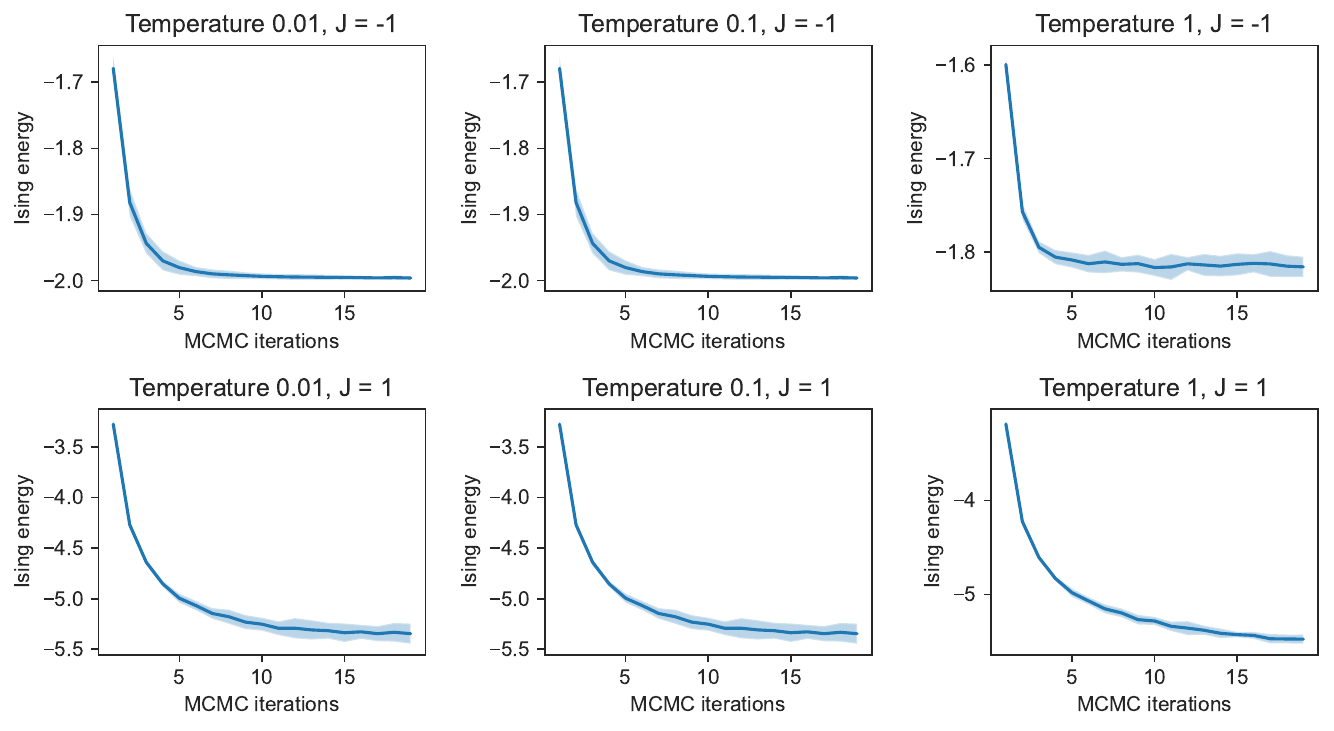}
    \caption{The mean and standard deviation of the Ising model energy against the number of Monte Carlo iterations for the ($J=1$) or antiferromagnetic ($J=-1$) Ising model. The simulation is also done for three different temperatures: 0.01, 0.1, and 1.}
    \label{fig:temp_ising}
\end{figure}

To illustrate the convergence rate of the Ising model, we plot the energy against the number of Monte Carlo iterations in \Figref{fig:temp_ising}. These simulations are done on the 20 meshes in the Bust dataset of Section~\ref{sec:mesh_sparse}. We plot the mean plus one standard dedication of the model energy at each iteration. We observe that the ferromagnetic ($J=1$) and antiferromagnetic ($J=-1$) Ising models converge fast, even though the antiferromagnetic one converges faster in just a handful of iterations. 

\newpage
\section{Graph Coloring}\label{app:graph_coloring}

\begin{figure}[ht]
    \centering
    \includegraphics[width = 0.7\linewidth]{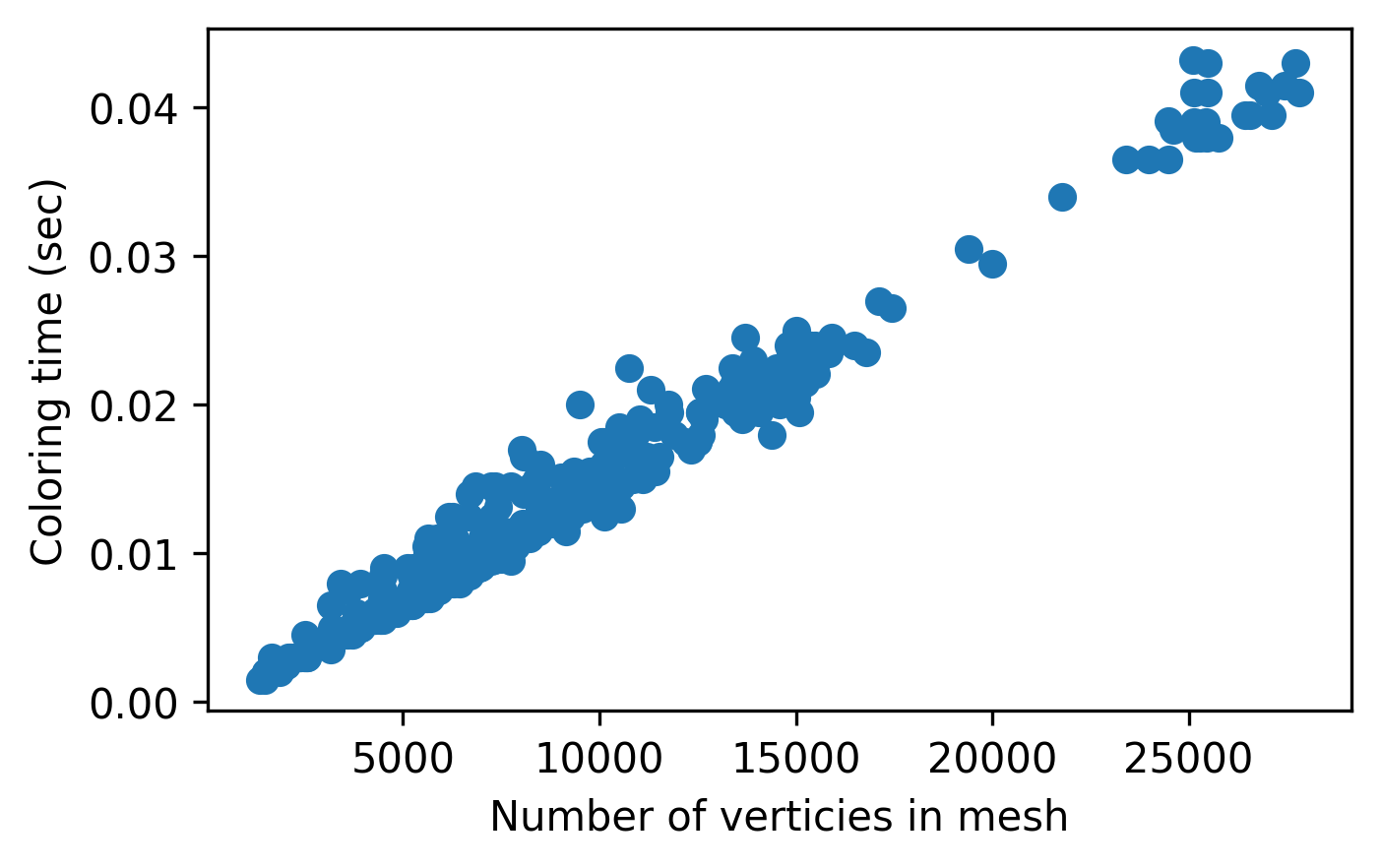}
    \caption{The time it takes to color a graph, plotted against the number of vertices in the graph.}
    \label{fig:coloring_time}
\end{figure}

The graph coloring is based on the greedy graph coloring algorithm explained in \cite{kosowski2004classical}. We use the implementation of NetworkX \cite{hagberg2008exploring}. The computational complexity of the greedy coloring method is $\mathcal{O}(m + n)$, where $n$ is the number of vertices and $m$ is the number of edges in the graph. Figure~\ref{fig:coloring_time} shows the coloring time plotted against the number of vertices for the graph coarsening dataset of Section~\ref{sec:mesh_sparse}. This indicates that the time taken for coloring is insignificant compared to the time required for sampling and coarsening. However, the coloring time could potentially become problematic for exceptionally large graphs. It is worth noting that coloring an image is extremely fast and straightforward due to its simple structure.

\newpage
\section{Binary Segmentation Details}\label{app:unet}
We train a U-Net for the bird segmentation task in Section~\ref{sec:segmentation} using three down-sampling blocks, one middle layer, and three up-sampling blocks. The implementation is inspired by \url{https://github.com/yassouali/pytorch-segmentation/tree/master}. The dimensions of the blocks/layers are presented in Table \ref{tab:unet_dim}.
\begin{table}[ht!]
\caption{Dimensions of the U-Net blocks.}
\label{tab:unet_dim}
\vskip 0.15in
\begin{center}
\begin{small}
\begin{sc}
\begin{tabular}{llll}
\toprule
Layer &  input dim &  output dim  &\\
\midrule
Input layer & 3 & 16   \\
Encoder 1 &16 & 32   \\
Encoder 2  & 32 & 64  \\
Encoder 3 & 64 & 128 \\
Mid layer & 128 & 128  \\
Decoder 1 &128 & 64   \\
Decoder 2  & 64 & 32  \\
Decoder 3 & 32 & 16 \\
Output layer & 16 & 1   \\
\bottomrule
\end{tabular}
\end{sc}
\end{small}
\end{center}
\vskip -0.1in
\end{table}

\begin{figure}[ht!]
    \centering
    \includegraphics[width=0.4\linewidth]{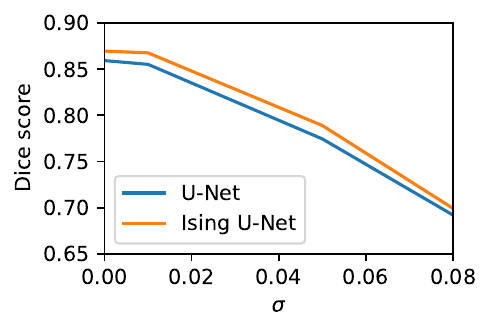}
    \caption{Dice score comparison of the two segmentation models under varying levels of noise, characterized by different standard deviations (\(\sigma\)), added to the input images.}
    \label{fig:sim_dice}
\end{figure}

We train the model using the Adam optimizer \citep{KingBa15}, a batch size of 32, and a learning range of $1e-4$. The dataset contains 6537 segmented images. We split the dataset in a train/val/test split using the ratios 0.8/0.1/0.1. We train the model for 300 epochs and test the model with the lowest validation error. Here, the Ising model uses a temperature of one and $J=1$. Figure~\ref{fig:sim_dice} illustrates that the Ising segmentation model is relatively robust to noise compared to using solely a U-Net.
\section{Explainability for Graph Classification Details}\label{app:graph_classsification_explainability}
This section contains the details of the explaining graph classification experiments of Section~\ref{sec:explain}.

\subsection{Datasets}
\begin{figure}[ht!]
    \centering
    \includegraphics[width=0.75\linewidth]{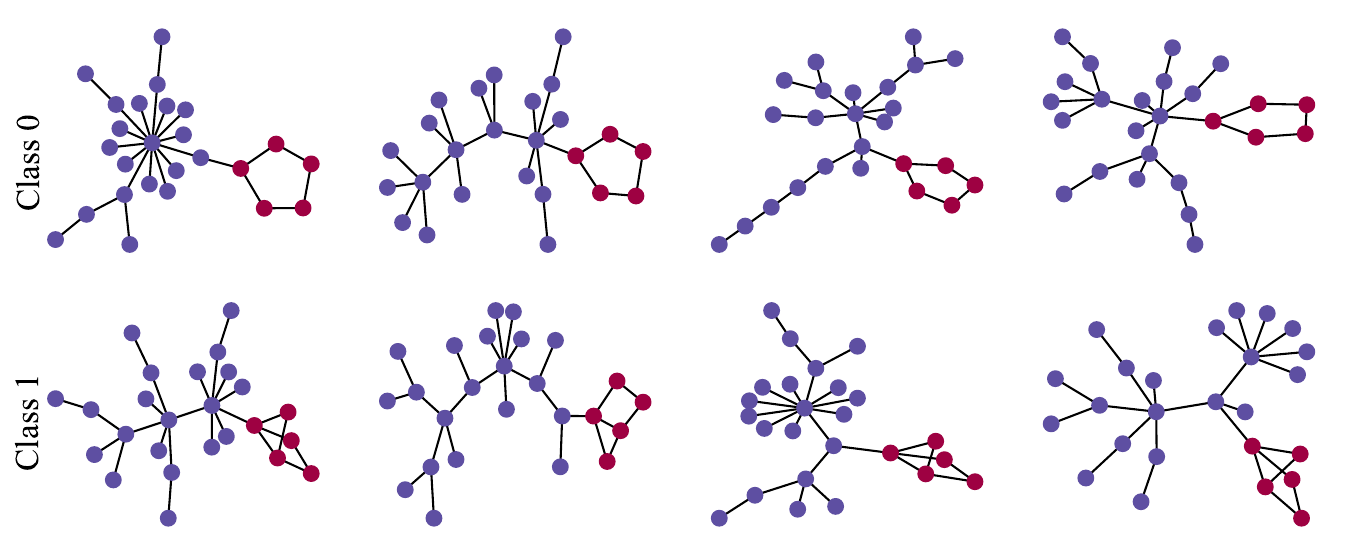}
    \caption{Samples from the two different classes in the BA-2motif dataset.}
    \label{fig:ba2_dataset}
\end{figure}

The \emph{BA-2motif} dataset, first presented in \cite{luo2020parameterized}, contains 1000 random Barabasi-Albert (BA) graphs where 50\% of the graphs have a house motif attached to it, while the other half has a cycling motif attached to it. The aim is to classify which of the two a graph belongs to. All graphs in the dataset contain 25 nodes. Figure~\ref{fig:ba2_dataset} depicts some samples.

\begin{figure}[ht!]
    \centering
    \includegraphics[width=0.99\linewidth]{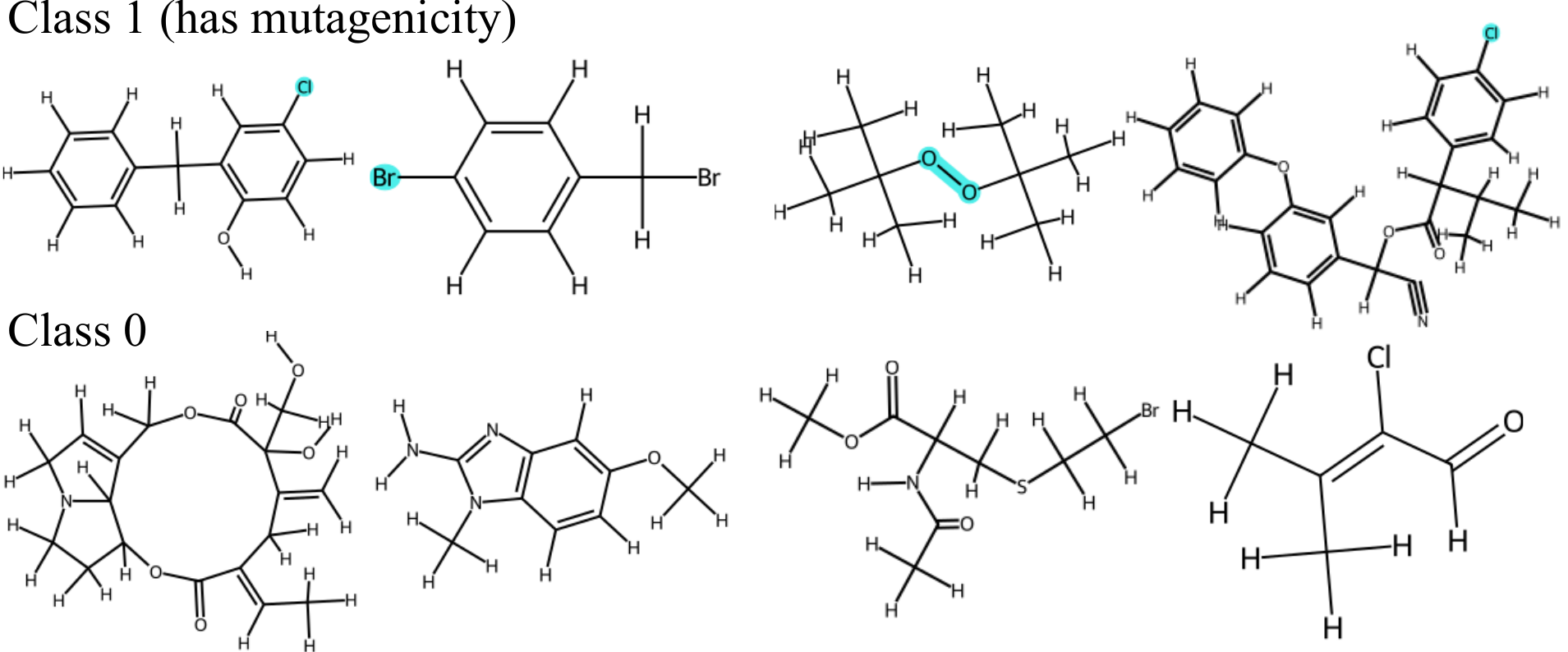}
    \caption{Samples from the two different classes in the Mutag dataset, with ground truth explanation labels highlighted for the positive (mutagenic) class. The figure shows that the provided explanations can be incomplete. For instance, chlorine (Cl) or bromine (Br) atoms attached to a carbon ring are associated with mutagenicity in the top row. However, in the bottom row, molecules containing Cl or Br are non-mutagenic, indicating that these atoms alone do not always lead to mutagenicity.}
    \label{fig:mutag_samples}
\end{figure}

The \emph{Mutag} (or \emph{Mutagenicity}) dataset \cite{kazius2005derivation} consists of 1,768 molecular graphs, categorized based on their mutagenic effects on the Gram-negative bacterium S. Although the original dataset contains 4,337 graphs, we utilize a curated version from \cite{agarwal2023evaluating}, which emphasizes the presence or absence of the toxicophores: NH$_2$, NO$_2$, aliphatic halides, nitroso, and azo groups. Upon reviewing the dataset and its ground truth explanations, we observe that they are occasionally incomplete. While the labels correctly identify the relevant atoms, they often fail to account for the atomic neighborhood. For example, as illustrated in Figure~\ref{fig:mutag_samples}, a chlorine (Cl) atom alone does not consistently result in a mutagenic molecule; the mutagenicity also depends on the atomic environment around the Cl atom. However, this contextual factor is not considered in the explanation labeling of the dataset. Consequently, the absolute value of the Graph Explanation Accuracy (GEA) metric should not be overemphasized; instead, it should be interpreted in relation to other models rather than as an isolated performance indicator.

\subsection{Baselines}\label{app:baslines_exp}
In our experiments, we evaluate six post-hoc explainability baselines: PGExplainer \cite{luo2020parameterized}, GNNExplainer \cite{ying2019gnnexplainer}, PGMExplainer \cite{pgmminh}, SubgraphX \cite{yuan2021explainability}, GStarX \cite{zhang2022gstarx}, and SAME \cite{NEURIPS2023_14cdc901}. These methods identify significant subgraph structures influencing GNN predictions. PGExplainer uses a neural network to learn patterns across multiple data points and identify subgraphs of importance, while GNNExplainer optimizes an edge mask for each instance using gradient descent. PGMExplainer utilizes probabilistic graphical models to capture node dependencies and offer interpretable explanations of their impact on predictions. SubgraphX applies Monte Carlo Tree Search with Shapley values to systematically explore subgraphs and generate globally interpretable explanations. GStarX leverages the HN value from cooperative game theory to efficiently identify influential subgraphs by considering structure-aware insights. SAME (Structure-Aware Model Explanation) explains GNNs by analyzing the contribution of each substructure while maintaining the balance of overall model complexity. We utilize the official PyTorch Geometric \cite{Fey/Lenssen/2019} implementations for PGExplainer, GNNExplainer, and PGMExplainer, the implementation from \cite{zhang2022gstarx} for GStarX and SubgraphX, and the original implementation for SAME. We follow the hyperparameters recommended by the authors. SAME and SubgraphX do not support edge features, so they are only evaluated on the BA-2Motifs dataset.

\subsection{Implementation Details}
For the explainability experiments in Section~\ref{sec:explain}, we use a three-layer GIN network \cite{xu2018powerful} with a hidden dimension of 64 for the BA-2Motifs dataset. For the Mutag dataset, we use a three-layer GINE network \cite{hu2019strategies}, which incorporates edge features. The hidden dimension is again 64. In all experiments, the classification model and the magnetic field model share the same structure, with the only difference being that a final mean pooling layer is applied to the graph classification model.

We train the classification models using the Adam optimizer \citep{KingBa15}, with a batch size of 64 and a learning rate of \(10^{-3}\), for a total of 500 epochs.

For our IGExplainer model, we use again the Adam optimizer \citep{KingBa15} with a batch size of 64 and a learning rate of \(10^{-3}\), and train for 300 epochs. The model with the lowest cross-entropy loss (see \Eqref{eq:crossent}) is saved. In these explainability experiments, we evaluate the model on the training set rather than using a separate test set. The focus of explainability is to interpret and understand the behavior of the trained model, not to assess its performance on new data.

\subsection{Additional Results}\label{app:ex_time}
\begin{figure}[ht!]
    \centering
    \includegraphics[width=0.7\linewidth]{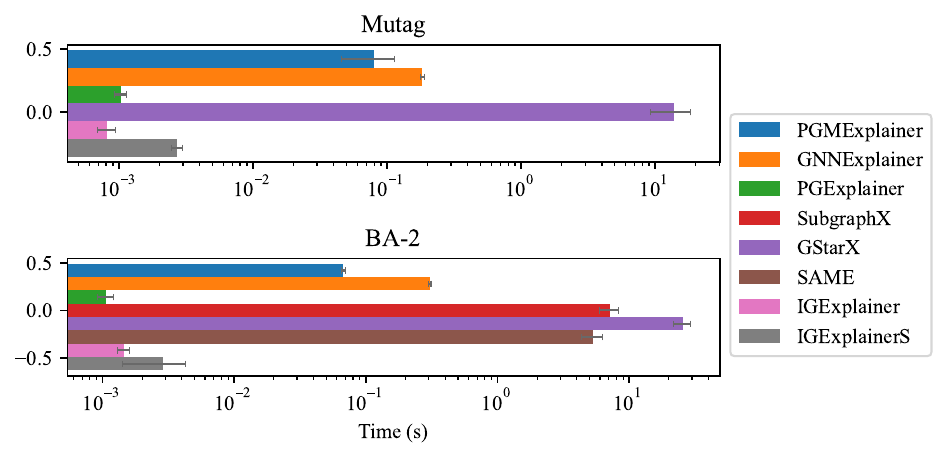}
    \caption{Inference time for all models used in the explainability experiments.}
    \label{fig:time_exp}
\end{figure}

The time required for each model to generate an explanation is shown in Figure~\ref{fig:time_exp}. IGExplainer and IGExplainerS are faster than all baselines except PGExplainer, which is the fastest model. IGExplainerS uses a few Monte Carlo samples to achieve the highest accuracy among the models, which makes it slightly slower than IGExplainer, which instead selects the top-k values of the magnetic field for its explanations. IGExplainerS, which provides the best performance, is still significantly faster (0.02 seconds on average for the BA-2Motif dataset) than GStarX, the slowest model, which takes an average of 25 seconds.

\begin{figure}[ht!]
    \centering
    \includegraphics[width=0.99\linewidth]{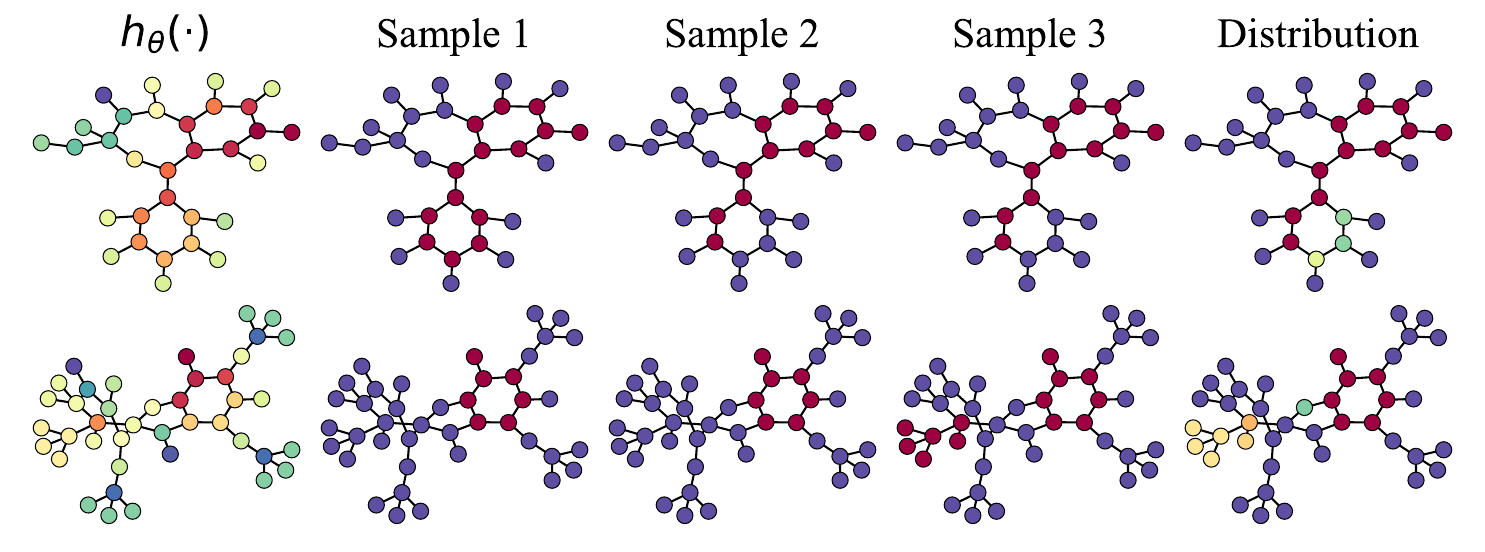}
    \caption{Samples from the trained IGExplainer. The distribution of the samples is different from the magnetic field $h_\theta(\cdot)$.}
    \label{fig:motifs_samples}
\end{figure}

When taking samples of the IGExplainer, every sample will differ since the Ising model is stochastic. This is visualized in Figure~\ref{fig:motifs_samples} for one graph in the \emph{BA-2motif} dataset. We also visualize several samples from the distribution and the sample average, i.e., approximately the marginal probability of including a node. Importantly, this differs from the magnetic field $h_\theta(\cdot)$.

\begin{figure}[ht!]
    \centering
    \includegraphics[width=0.99\linewidth]{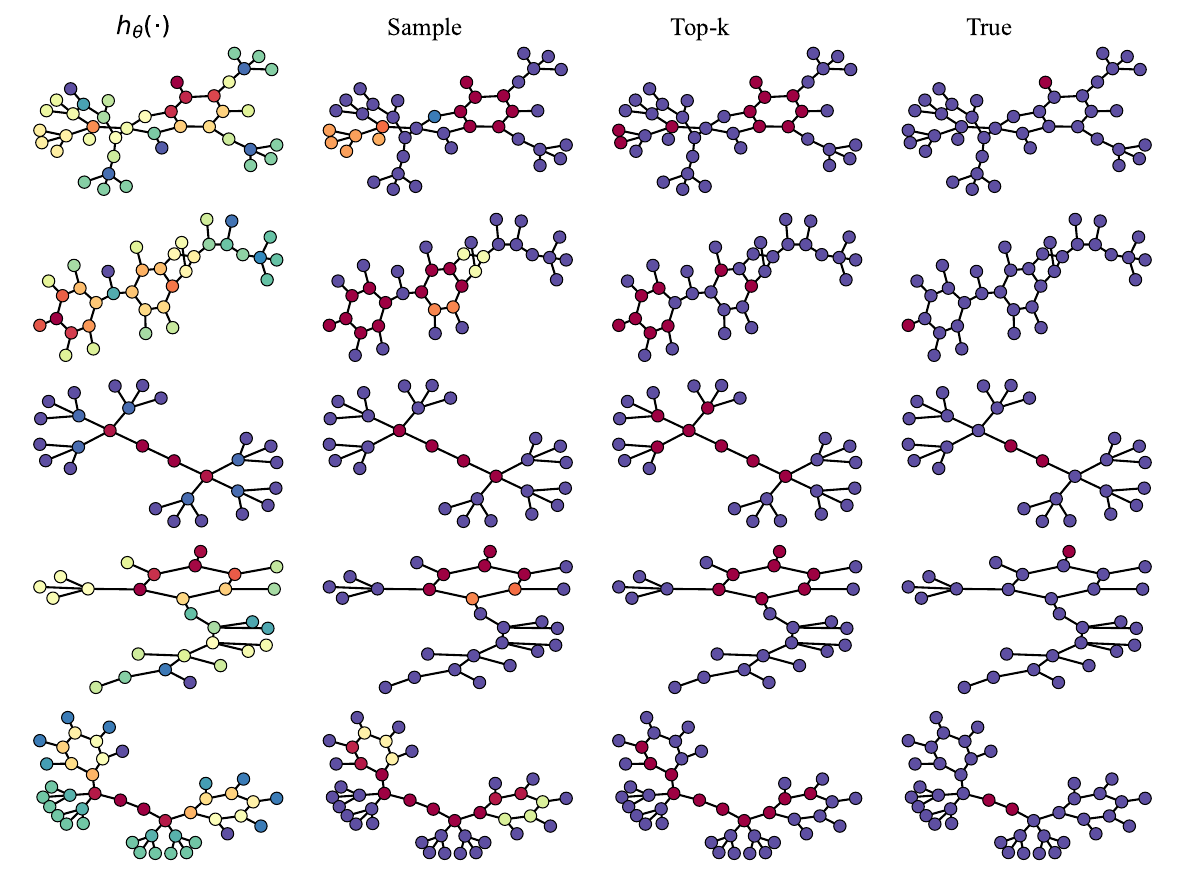}
    \caption{Output of the IGExplainer on the Mutag dataset. From left to right: the predicted magnetic field \( h_\theta(\cdot) \), the sample distribution, the top-25\% of nodes where \( h_\theta(\cdot) \) is the score, and finally the ground truth.}
    \label{fig:mutag_multioutput}
\end{figure}

Additional predictions from the IGExplaner on the Mutag dataset are presented in Figure~\ref{fig:mutag_multioutput}.
\clearpage

\section{Mesh Simplification}\label{app:mesh_sparse}
This section contains extra information on the mesh simplification experiments of Section~\ref{sec:mesh_sparse}. 

\subsection{Dataset}
The details of the mesh dataset are presented in Table~\ref{tab:mesh_dataset_info}.

\begin{table}[ht!]
\caption{Details of the meshes in the dataset used in Section~\ref{sec:mesh_sparse}. The table presents the mean number and the standard deviation in parentheses.}
\label{tab:mesh_dataset_info}
\vskip 0.15in
\begin{center}
\begin{small}
\begin{sc}
\begin{tabular}{lcccc}
\toprule
dataset & Bust & Fourleg & human & chair \\
\midrule
Number of vertices &  25662 (1230) & 9778 (3671)  & 8036 (4149) & 10154 (999) \\
Number of faces &51321 (2460) &19553 (7342) & 16070 (8302) & 20313 (2002) \\
\bottomrule
\end{tabular}
\end{sc}
\end{small}
\end{center}
\vskip -0.1in
\end{table}

\subsection{Implementation Details}\label{app:eqnet}
The equivariant GNN is constructed using the \emph{e3nn} library \citep{geiger2022e3nn}. As the node input, we derive the mesh Laplacian according to
\begin{equation}
\mathbf{C}_{ij} =  \begin{cases}
            w_{ij} = \frac{\cot \alpha_{ij} + \cot \beta_{ij}}{2}, &
            \text{if } i, j \text{ is an edge}, \\
            -\sum_{j \in N(i)}{w_{ij}}, &
            \text{if } i \text{ is in the diagonal}, \\
            0, & \text{otherwise},
        \end{cases}
\end{equation}
where and $\alpha_{ij}$ and $\beta_{ij}$ denote the two angles opposite of edge $(i, j)$ \citep{sorkine2005laplacian}. We get a measure of the  curvature of the node using 
\begin{equation}
    z_i^0 = \left \| \sum_{i \in \mathcal{N}(i)}\mathbf{C}_{ij} (\hat{r}_i- \hat{r}_j) \right \|^2_2,
\end{equation}
where $\hat{r}_i$ is the position of node $i$. We use $z_i^0$ as the model node input. As edge attributes, we use the spherical harmonics expansion of the relative distance between node $i$ and its neighbors $j$, $Y\left ((\hat{r}_i - \hat{r}_j)/\| \hat{r}_i - \hat{r}_j\|) \right)$ and the node-to-node distance $d_{ij} = \| \hat{r}_i - \hat{r}_j \|$. We construct an equivariant convolution by first deriving the message from the neighborhood as
\begin{equation}
    \hat{z}^{k+1}_i = \frac{1}{n_d} \sum_{j\in \mathcal{N}(i)} z_j^k \otimes_{f_{W}(d_ij)}Y\left (\frac{\hat{r}_i - \hat{r}_j}{\| \hat{r}_i - \hat{r}_j \|}\right),
\end{equation}
where $n_d$ is the mean number of neighboring nodes of the graph. For a triangular mesh, the degree is commonly six. The operator $\otimes_{f_{W}(d_ij)}$ defines the tensor product with distant dependent weights $f_{W}(d_ij)$; this is a two-layer neural network with learnable weights $w$ that derives the weights of the tensor product. The node features are then updated according to
\begin{equation}
    {z}^{k+1}_i = z^k_i + \alpha \cdot \hat{z}^{k+1}_i,
\end{equation}
where $\alpha$ is a learnable residual parameter dependent on $z^k_i$. We took inspiration for the model from the e3nn homepage, see
\url{https://github.com/e3nn/e3nn/blob/main/e3nn/nn/models/v2106/points_convolution.py}.

We construct a two-layer equivariant network where the layer irreducible representations are [0e, 16x0e + 16x1o, 16x0e]. This means that the input (0e) consists of a vector of 16 scalars, and the hidden layers consist of vectors with 16 scalars (16x0e) together with 16 vectors (16x1o) of odd parity. This, in turn, maps to the output vector of 16 scalars (16x0e). We subtract the representation with the mean of all nodes in the graph and finally map this representation to a single scalar output. Here, we use a temperature of one and $J=-1$ for the Ising model.  

For optimization, we employ the Adam optimizer \citep{KingBa15} with a learning rate of 0.001 and a batch size of one. We train the model until convergence on the training set but with a maximum of 50 epochs. We did not use a validation set as there were no indications of overfitting.

\subsection{Baselines}
We compare our {ISING+MAG} model to five other sampling methods. 
\begin{itemize}
    \item {RANDOM} is regular random sampling, where every node has a 50\% chance of being sampled, and the samples are independent.
    \item  {FPS} stands for farthest point sampling, also sometimes called farthest point strategy. FPS selects a subset of points from a point cloud by iteratively picking the point farthest from the already chosen points, maximizing the coverage of the original set.
    \item {POINT SAMPLER} is the point sampler in \cite{potamias2022neural}, which uses a traditional GNN approach. It aimed to be faster than the iterative FPS and better than random sampling by using a GNN with a specific \emph{DevConv} layer to evaluate each point's importance by analyzing its deviation from neighboring points.
    \item {SPECTRAL} is based on spectral coarsening \cite{shuman2015multiscale}, which uses the polarity of the largest Laplacian eigenvector, which, just as the Ising models, aims to find an ``every other'' pattern.
    \item {ISING} is the Ising model, but where the magnetic field is set to zero, resulting in a model that aims to sample in an ``every other'' pattern.
\end{itemize}
These methods cannot be trained for specific downstream tasks; they are inherently designed to distribute samples across the mesh evenly rather than optimize for particular applications.
\subsection{Training Time}
Table~\ref{tab:mesh_traning_time} presents the ISING+MAG and POINT SAMPLER training times. The extended training time for ISING+MAG is primarily due to the need to coarsen the mesh to compute the loss.

\begin{table}[ht!]
\caption{Training time per epoch (in seconds) for the mesh sparsification problem. We train the ISING+MAG model for 40-60 epochs and the point sampler for 150. The largest part of the time comes from redrawing the edges in the coarse mesh.}
\label{tab:mesh_traning_time}
\begin{small}
\begin{sc}
\begin{center}
\begin{tabular}{lcccc}
    \toprule 
    \multicolumn{1}{l}{DATASET} &
    \multicolumn{1}{c}{Bust} &
    \multicolumn{1}{c}{Fourleg} &
    \multicolumn{1}{c}{Human} &
    \multicolumn{1}{c}{Chair} \\ 
    \midrule
ISING+MAG& 35.22 & 9.32 & 11.82 & 12.99|
 \\
POINT SAMPLER &0.48& 0.174& 0.162  &0.213  \\
\bottomrule
\end{tabular}
\end{center}
\end{sc}
\end{small}
\vskip -0.1in
\end{table}

\subsection{Extra Results}
The outcome of the experiments presented as a bar chart in Figure~\ref{fig:dog_points} is presented in Table~\ref{tab:results_mesh_sampling_extra}.

\begin{table}[ht!]
\caption{Mean and standard deviation of the test vertex-to-mesh distance from the five-fold cross-validation of the four datasets. Lower is better.}
\label{tab:results_mesh_sampling_extra}
\begin{small}
\begin{sc}
\begin{center}
\begin{tabular}{lcccc}
    \toprule 
    \multicolumn{1}{l}{DATASET} &
    \multicolumn{1}{c}{Bust} &
    \multicolumn{1}{c}{Fourleg} &
    \multicolumn{1}{c}{Human} &
    \multicolumn{1}{c}{Chair} \\ 
    \midrule
ISING+MAG& \textbf{0.221} (0.042) &\textbf{0.339} (0.034) &\textbf{0.366} (0.031)  &\textbf{0.369} (0.046) \\
ISING & 0.288 (0.055)  &0.411 (0.034)  &0.409 (0.033)  &0.548 (0.045) \\
RANDOM & 0.343 (0.064)  & 0.498 (0.032)& 0.498 (0.032)  &0.592 (0.040)\\
SPECTRAL & 0.290 (0.056)  & 0.415 (0.034) & 0.412 (0.034)  &0.569 (0.054)\\
POINT SAMPLER &0.298  (0.057) & 0.476  (0.034)& 0.494  (0.034)  & 0.576 (0.046) \\
FPS &0.303  (0.059)& 0.395  (0.029) & 0.417 (0.015)  & 0.619 (0.059) \\
\bottomrule
\end{tabular}
\end{center}
\end{sc}
\end{small}
\end{table}
\vspace{0.3in}

The results presented in Table~\ref{tab:results_mesh_sampling_extra} show that the Ising model with the magnetic field (ISING+MAG) surpasses all other models regarding the point-to-mesh distance. This outcome aligns with expectations, given that our model is trained for this specific task. Their results are virtually identical when comparing the Ising model without the magnetic field (ISING) to Spectral coarsening (SPECTRAL). This is anticipated since both approaches result in an every-other pattern. In SPECTRAL, nodes are chosen based on the polarity of the largest eigenvector of the graph Laplacian, which divides the nodes into two equally large subgroups.

While the farthest point sampling (FPS) method ensures that samples are distributed across the graph, it has its limitations. In some datasets, it may outperform ISING and SPECTRAL. However, it is important to note that FPS does not guarantee an ``every other'' pattern and may inadvertently miss nodes in critical areas of the mesh.

Overall, the Ising model with the magnetic field (ISING+MAG) outperforms all other models. Similar to ISING and SPECTRAL, it samples an ``every other'' pattern. However, leveraging a geometry-aware network also learns to identify nodes crucial for maintaining mesh shape. This behavior, which defaults to using the ``every other'' pattern and selectively samples nodes based on the geometric shape of surrounding nodes, contributes to the model's effectiveness.

\begin{figure}[ht!]
    \centering
    \includegraphics[width = 0.6
    \linewidth]{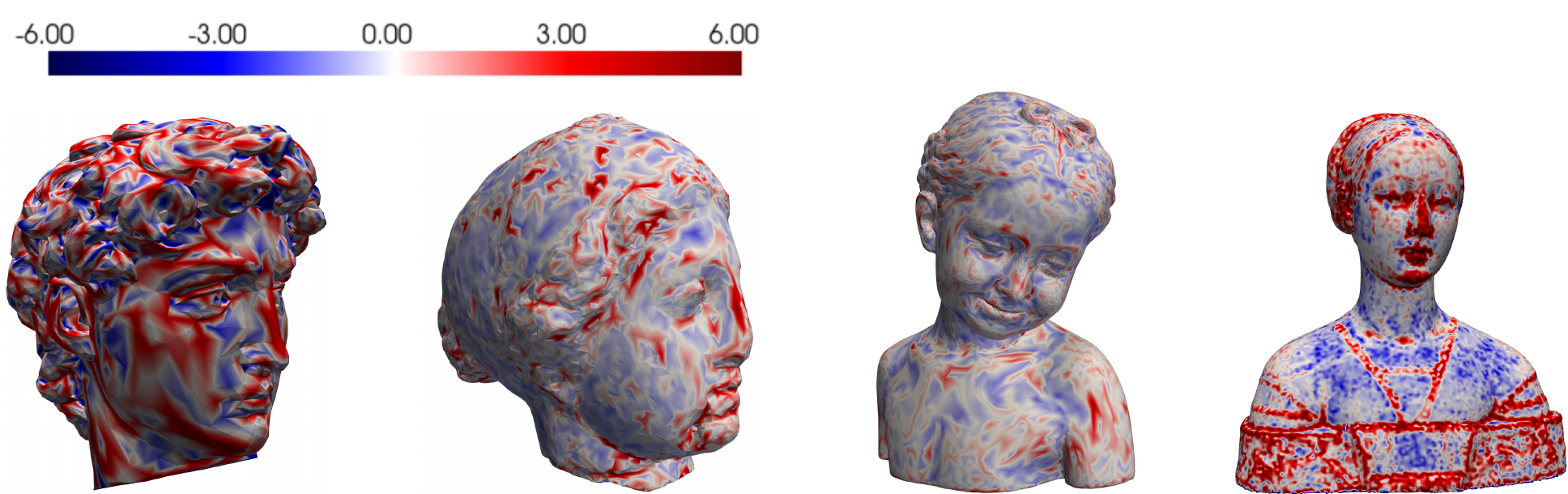}
    \caption{Examples of the magnetic field from the bust dataset (test set).}
    \label{fig:mag_bust}
\end{figure}

\begin{figure}[ht!]
    \centering
    \includegraphics[width = 0.6
    \linewidth]{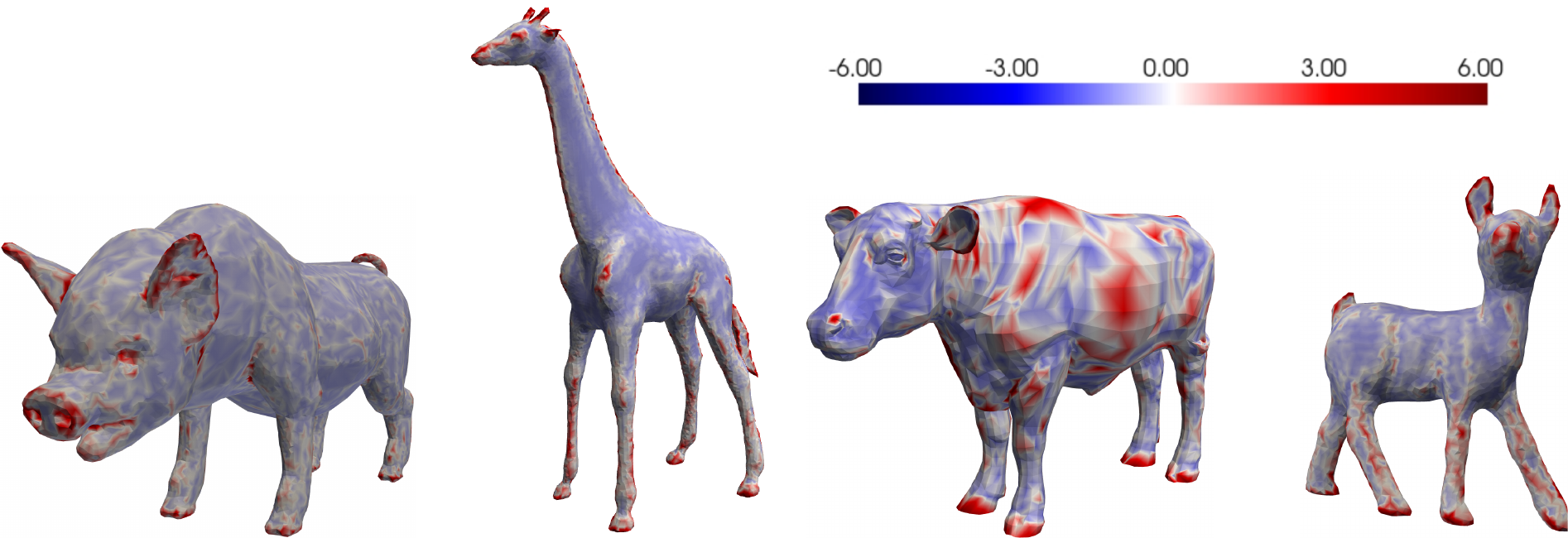}
    \caption{Examples of the magnetic field from the fourleg dataset (test set).}
    \label{fig:mag_fourleg}
\end{figure}

\begin{figure}[ht!]
    \centering
    \includegraphics[width = 0.6
    \linewidth]{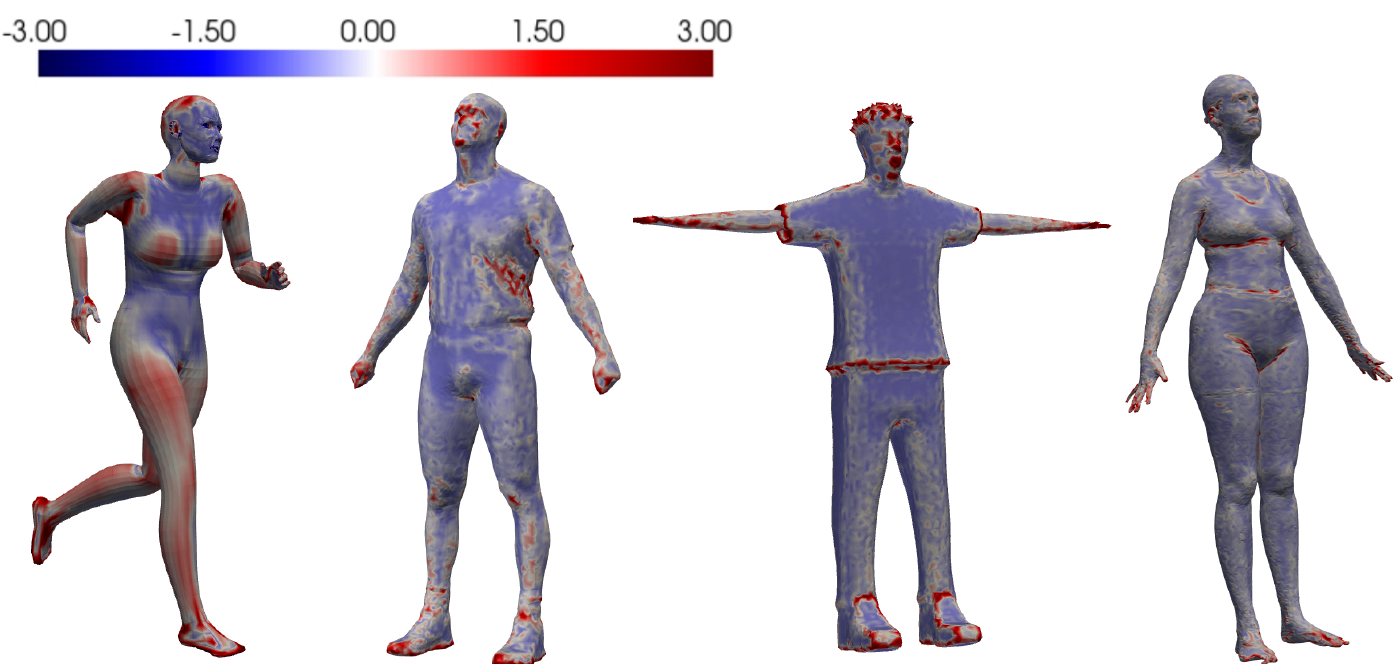}
    \caption{Examples of the magnetic field from the human dataset (test set).}
    \label{fig:mag_human}
\end{figure}

\begin{figure}[ht!]
    \centering
    \includegraphics[width = 0.6
    \linewidth]{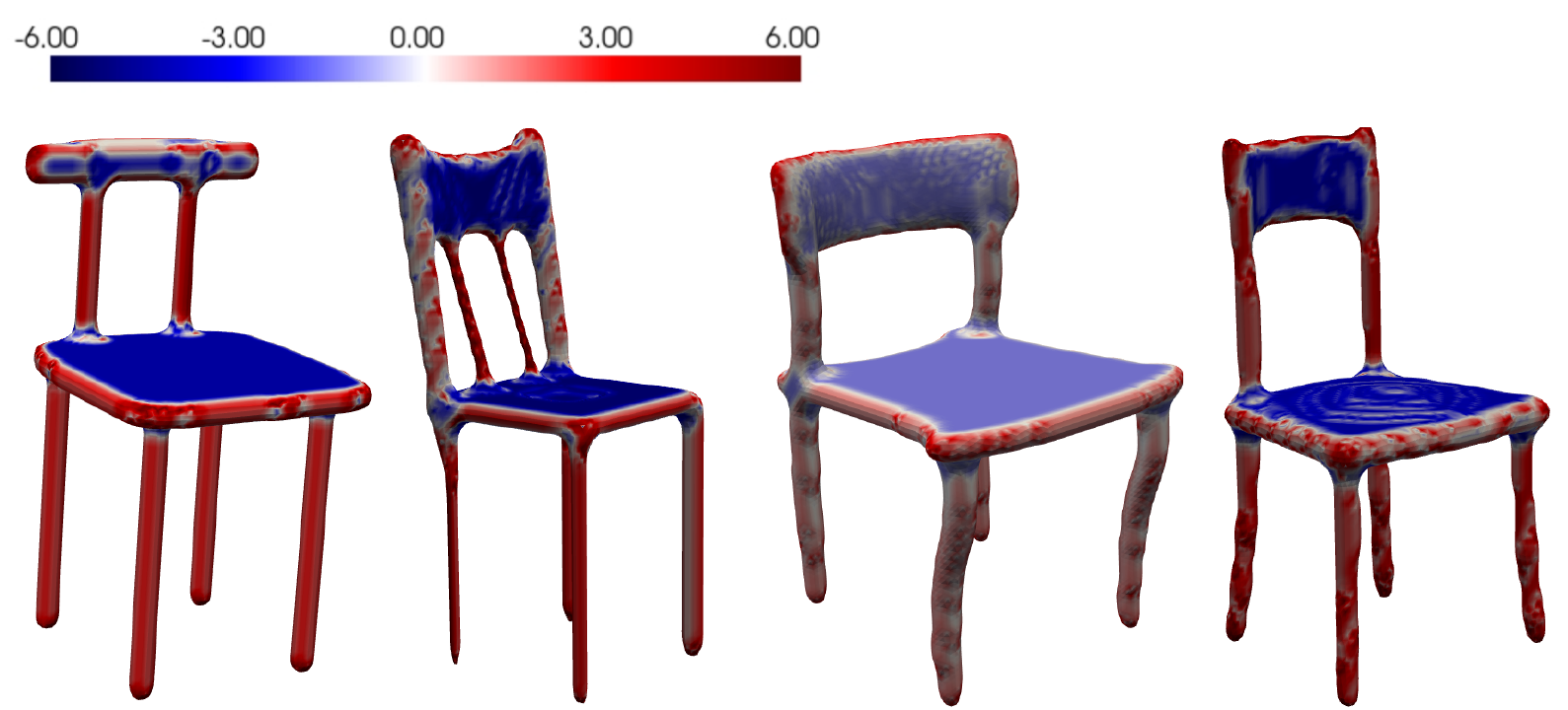}
    \caption{Examples of the magnetic field from the chair dataset (test set).}
    \label{fig:mag_chair}
\end{figure}

\begin{figure}[ht!]
    \centering
    \includegraphics[width = 0.9
    \linewidth]{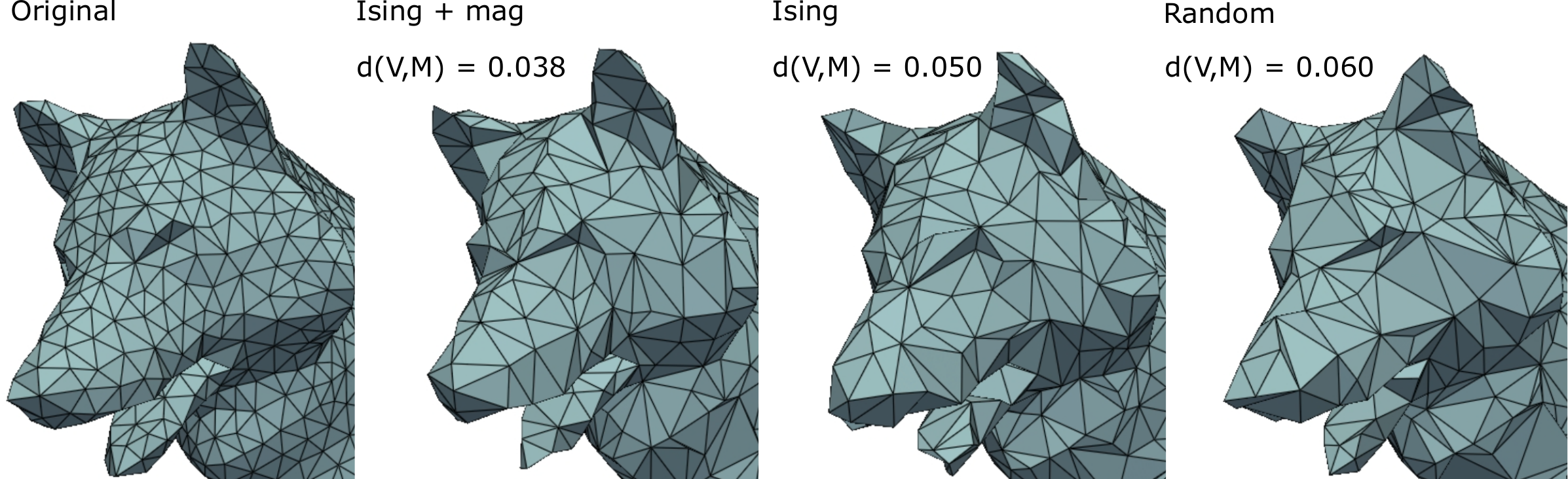}
    \caption{The coarsened mesh using random subsampling, Ising subsampling, and learned Ising subsampling of the nodes. The point-to-mesh distance is reduced from 0.06 to 0.038 when going from the random subsampling to the learned Ising graph. The time it takes to sample and create the course mesh is random, 1.37 sec, Ising 0.67 sec, and Ising+mag 1.31 sec. Random sampling takes longer due to a more complicated course mesh creation.}
    \label{fig:deg_deff_meshes}
\end{figure}

Examples of how the Ising model learns important areas of the mesh are presented in  Figures~\ref{fig:mag_bust}, \ref{fig:mag_fourleg}, \ref{fig:mag_human}, and \ref{fig:mag_chair}. Here, we visualize the magnetic field from the test split from the four datasets. An illustration of the course meshes after sampling is presented in Figure~\ref{fig:deg_deff_meshes}. 


\begin{figure}[ht!]
    \centering
    \includegraphics[width = 0.95\linewidth]{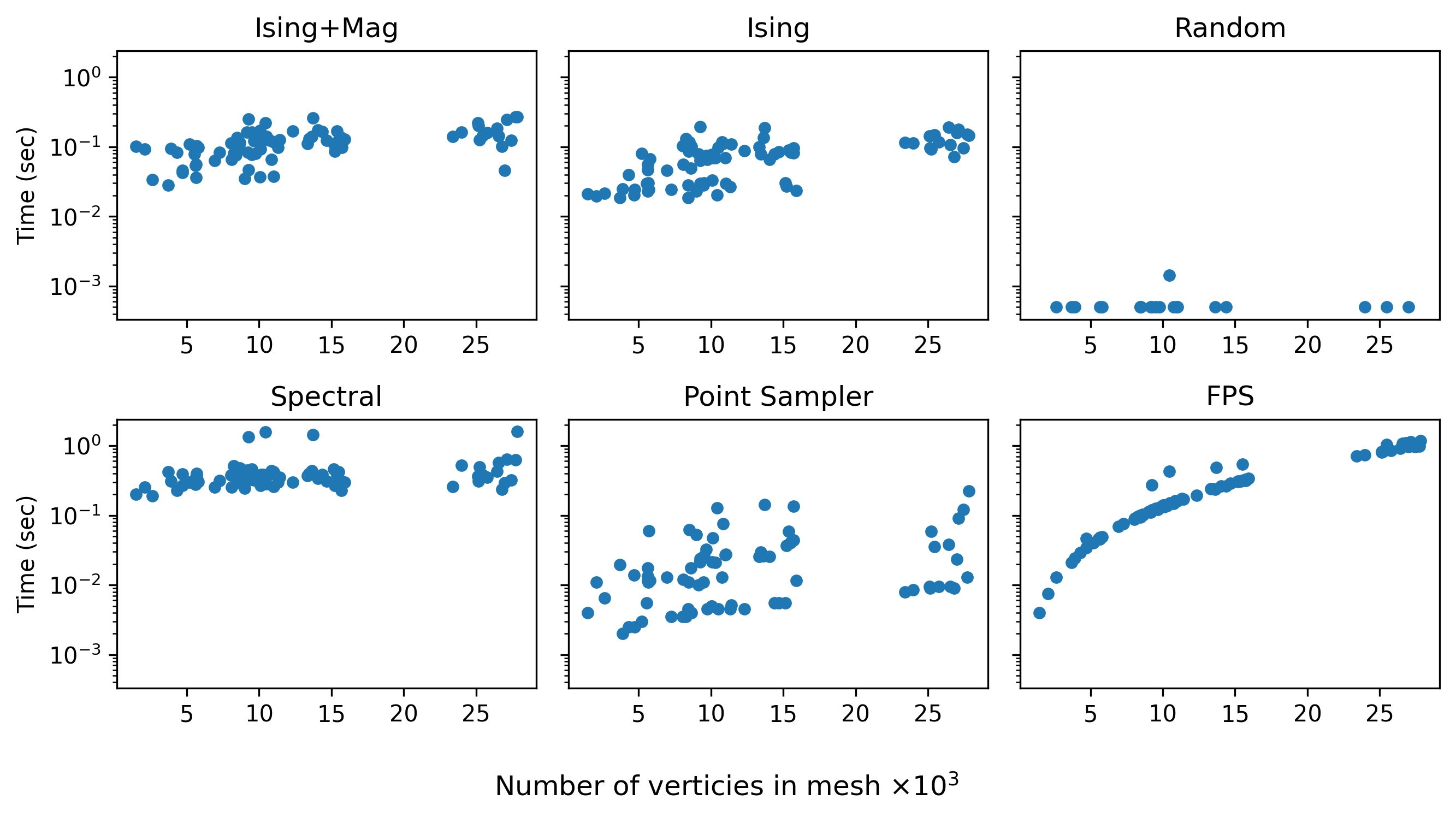}
    \caption{The time it takes to get sample nodes to include in the coarse graph.}
    \label{fig:sim_time_only}
\end{figure}

\newpage
\myparagraph{Complexity.} Given the large graphs in mesh processing, evaluating time complexity is crucial. We measure the time complexity for coarsening and sampling to complement the results in Table~\ref{tab:results_mesh_sampling_extra}. Figure~\ref{fig:sim_time_only} shows the sampling time alone, excluding mesh reconstruction, as a function of the mesh size. Notably, only random sampling and the neural network-based point sampler are faster than the Ising model. The relatively efficient scaling of the Ising model is due to two factors: (i) the parallelism enabled by the Metropolis-Hastings algorithm through graph coloring, and (ii) the fast convergence of the Ising model, which requires only a few iterations of the Metropolis-Hastings algorithm.

\begin{figure}[ht!]
    \centering
    \includegraphics[width = 0.95\linewidth]{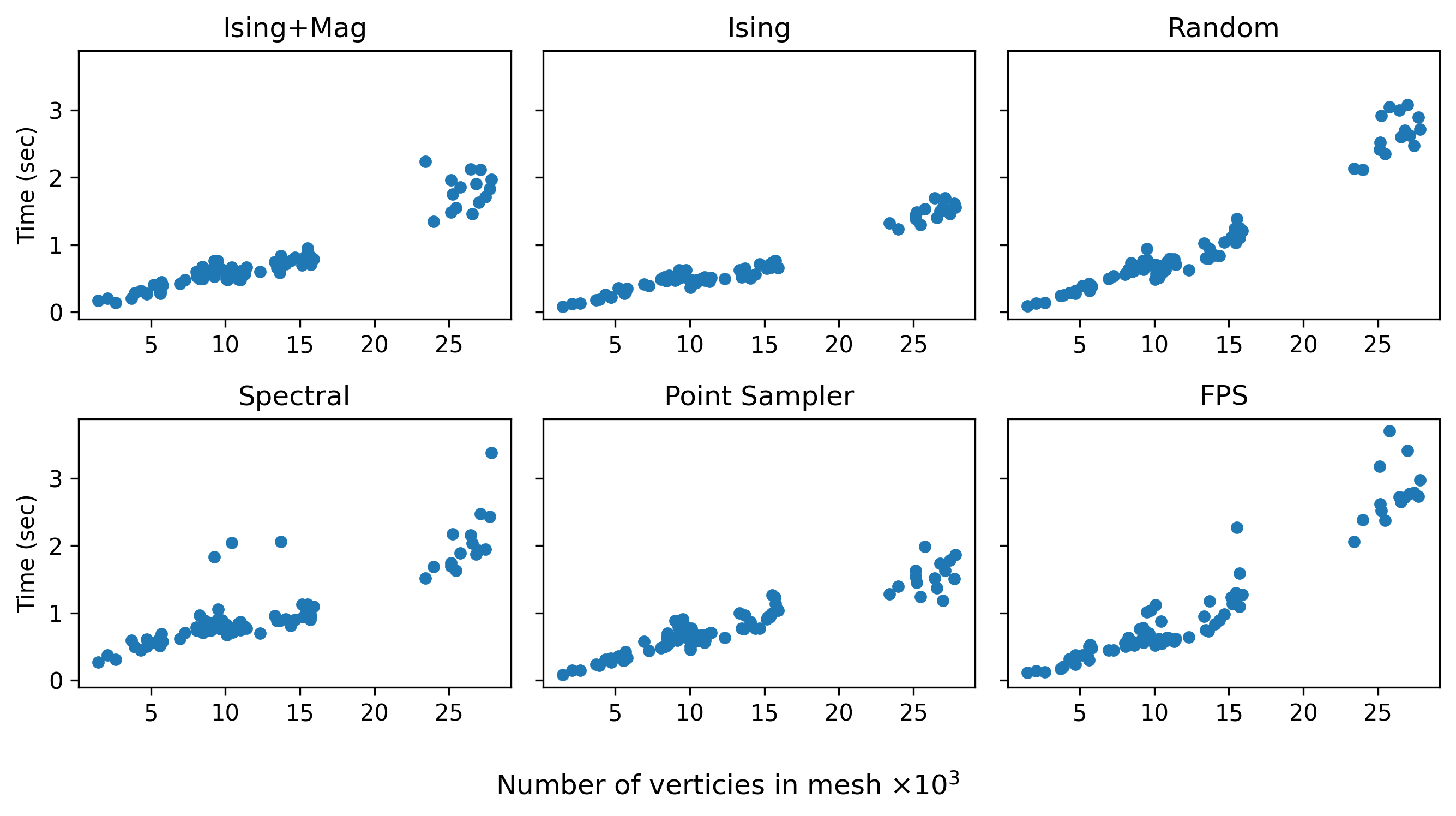}
    \caption{The time it takes to sample and coarsen the graph, plotted against the number of vertices in the graph.}
    \label{fig:sim_timey}
\end{figure}

Figure \ref{fig:sim_timey} illustrates the combined time for sampling and coarsening. It shows that the Ising model's coarsening time is lower than that of random sampling. This is because coarsening is less efficient with unevenly distributed samples. Additionally, coarsening is the major time factor, while the sampling time is negligible.

\clearpage
\section{Sparse Approximate Matrix Inverses}\label{app:matrix_appendix}

\subsection{Results and Implementation Details} \label{app:matrix_results}

With our initial model architecture, the graph representation induces an a priori assumption on the sparsity pattern of the predicted sparse approximate inverse in which we aim to select $50\%$ of the elements in the sparsity pattern of $A+A^2$. Our approach is, however, not limited to the choice of an a priori sparsity pattern. Implementation-wise, we lift this constraint by adding an extra edge feature consisting of an all-ones $n\times n$ matrix in the graph representation of the input matrix, allowing for a selection of $50\%$ of all matrix elements in the sparse approximation of the inverse. As the a priori sparsity pattern is already incorporated into the graph representation of the input matrix, this renders our method flexible in terms of choice of a priori sparsity pattern, which in our case should be denser than the desired sparsity level. This flexibility is expected to be especially important for computational efficiency in use cases involving large matrix dimensions.

Given the current state of the trainable parameters, two samples from the Ising model are required during training to obtain the gradient estimate. Evaluation is carried out using a single sample. In both cases, the sample $x$ from the Ising model produced after $T$ MCMC iterations is passed to the final layer, which solves the the $n$ optimization problems, 
\begin{equation} 
 \begin{aligned}
 s^*_k = \argmin_{s'_k \in \mathcal{R}^{m_k}} {\lVert AM_k(s_k) - I_k \rVert}^2_2,
 \end{aligned}
 \end{equation}
computing the values $s^*_k \in \sR^m$ of the non-zero entries of each column $M_k$ of $M$. Thus, the final output is the predicted sparse approximate inverse $M$. Empirically, $T=3$ is found to be sufficient.

Independent on which dataset is used, $60\%$ of the samples are used for training, and the remaining $40\%$ are divided equally between a validation set and a test set. We use the magnetic field-dependent regularization described in Section~\ref{sec:sampling_fraction} to optimize for a sampling fraction of $50\%$ of the a priori sparsity pattern. In all settings, the sampling fraction converges to an average value close to $50\%$ with the proposed regularization scheme. The mean recorded sampling fraction on the test dataset in \emph{Setting 1} and \emph{Setting 2} is $52.4\%$ and $50.9\%$, respectively. The corresponding sampling fraction in \emph{Setting 3} is $49.8\%$. The baseline methods are tuned to allow for the same sampling fraction during evaluation.

Results for two samples from the test dataset in \emph{Setting 3} are visualized in Figure~\ref{fig:matrix_226_58}. The first sample (top) shows the predicted sparsity pattern when the true inverse is dense. For this sample, the recorded loss is $2.61$ (ISING+MAG) compared to $3.57$ (Ising), $4.18$ (Random) and $4.12$ (Only~A). As discussed in Figure~\ref{fig:sampling_external_magnetic}, the magnitude of the local magnetic field influences the sampling probability in that region, which can be seen by comparing the output magnetic field to the obtained sparsity pattern for both samples. A zero magnetic field in the Ising model can be interpreted as a prior on the sparsity pattern in which $50\%$ of the elements in each row and column of the matrix are selected, resulting in an inverse approximation in which the nonzero elements are evenly distributed across the given matrix. For the second sample (bottom), the sparsity optimized for is lower than the sparsity of the true inverse. We choose to display this sample as the results clearly illustrate that the learned magnetic field indeed results in an Ising model from which the sampled sparsity pattern successfully avoids the most unimportant regions in the true inverse for the particular input matrix. For this sample, the recorded loss is $1.02$ (ISING+MAG) compared to $3.75$ (Ising), $3.76$ (Random) and $4.06$ (Only~A).

The current baselines include comparisons to uniform random sampling and an Ising model without the learned magnetic field key to our approach, both restricted to the allowed sampling fraction. This restriction is key to fair comparisons since including more nonzeros can improve the performance metric. With this in mind, comparison to assuming a sparsity pattern corresponding to the sparsity pattern of input matrix A is a good sanity check, but in this case worse performance can be expected if the sparsity of the predicted approximate inverse is lower.

\subsection{Graph Convolutional Neural Network }\label{app:gcnnet}
Here, we use the simple and deep graph convolutional networks from \citet{chen2020simple} where we set the strength of the initial residual connection $\alpha$ to $0.1$ and $\theta$, the hyperparameter to compute the strength of the identity, to $0.5$. The network contains four layers, each with a hidden dimension of 64. We utilize weight sharing across these layers. The dimension of the output of the final convolutional layer is 64. This output is subtracted with the mean of the node values before a final linear layer reduces the size to one. 

We employ the Adam optimizer \citep{KingBa15} with a learning rate of 0.01 for optimization. We train the model until convergence on the training set but with a maximum number of 300 epochs, as there were no indications of overfitting. Here, the Ising model uses a temperature of one and $J=-0.4$.  

The training times per epoch were 1 minute for \emph{Setting~1} and three minutes for \emph{Setting~2} and \emph{Setting~3}, respectively. These models converged in less than 50 epochs.
\newpage
\subsection{Datasets}\label{app:datasets}
\begin{figure*}[ht]
    \centering
    \includegraphics[width = 0.75\linewidth]{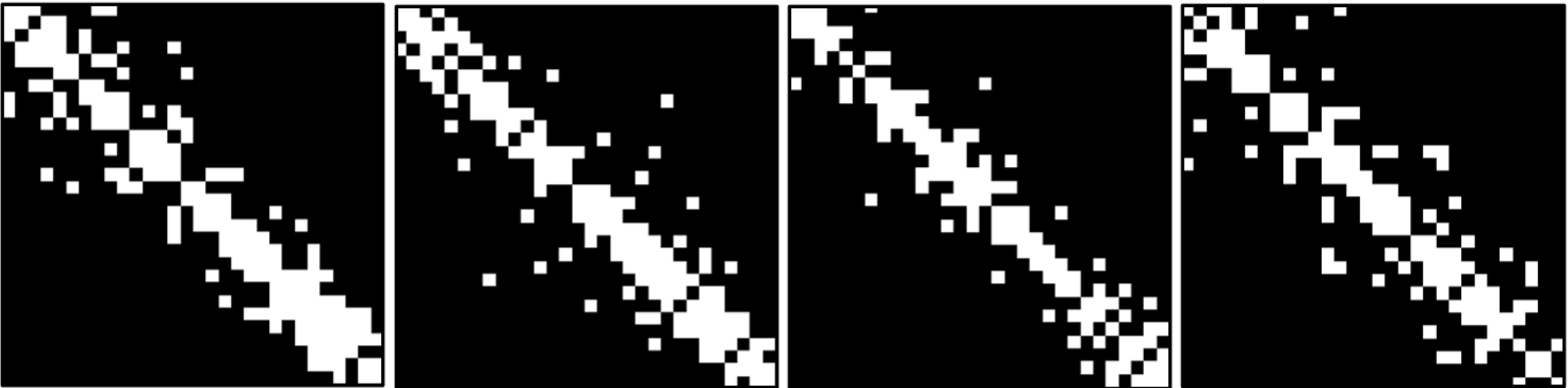}
    \caption{Example of samples from \emph{Dataset 1}. Nonzero elements are represented in white.} 
    \label{fig:dataset1}
\end{figure*}

\myparagraph{Dataset~1.}
This is a synthetic dataset containing 1600 binary matrices. Each matrix is real, symmetric, and of dimensionality $30\times30$. The matrices in the dataset are automatically generated based on the principle that the probability of a nonzero element increases towards the diagonal, with $100\%$ probability of nonzero diagonal elements. The upper right and lower left corners have a very low probability of being nonzero, and in between these elements and the diagonal, the probability of a nonzero element varies according to a nonlinear function. The mean sparsity is $83\%$ zero-elements in the generated dataset. The maximum sparsity allowed in the data-generating process is $96\%$, and the determinant of each matrix is bounded between $0.001$ and $50$. Figure \ref{fig:dataset1} shows sparsity patterns of four matrices in the final dataset.

The dataset is available upon request and will be released together with the code used for preprocessing and generation.

\begin{figure*}[ht]
    \centering
    \includegraphics[width = 0.75\linewidth]{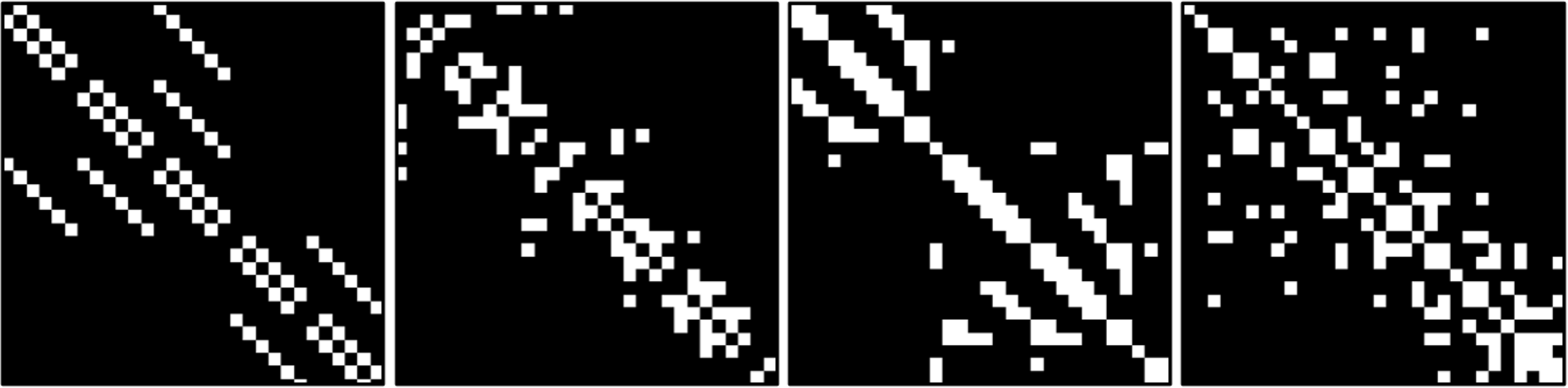}
    \caption{Example of sparsity patterns of samples from \emph{Dataset 2}. Nonzero elements are represented in white.} 
    \label{fig:dataset2}
\end{figure*}

\myparagraph{Dataset 2.}
This is a synthetic dataset containing 1800 submatrices of size $30\times30$ constructed from the SuiteSparse Matrix Collection \cite{matrixcollectionref}. Since the original matrices in the dataset are often denser closer to the diagonal, the submatrices are constructed by iterating a window of size $30\times30$ along the diagonals, such that no submatrix overlaps another submatrix. To ensure that each submatrix is symmetric, only the upper triangular part of the matrix is used to create each matrix. Each submatrix is scaled such that the absolute value of the maximum element is one, and submatrices with all values below $10^{-6}$ are removed from the dataset. The mean sparsity in the final dataset is $89\%$. The maximum sparsity allowed in the data-generating process is $96\%$, and the determinant of each matrix is bounded between $0.001$ and $50$. Figure~\ref{fig:dataset2} shows sparsity patterns of four matrices in the final dataset.

The original dataset is under the  CC-BY 4.0 License. Our modified dataset is available upon request and will be released with the code used for preprocessing and generation.

\end{document}

%% file: math_commands.tex
\usepackage{amsmath,amsfonts,bm}

\usepackage{letltxmacro}
\LetLtxMacro{\originaleqref}{\eqref}

\def\Figref#1{Figure~\ref{#1}}

\def\eqref#1{equation~\originaleqref{#1}}
\def\Eqref#1{Equation~\originaleqref{#1}}

\def\1{\bm{1}}

\DeclareMathAlphabet{\mathsfit}{\encodingdefault}{\sfdefault}{m}{sl}
\SetMathAlphabet{\mathsfit}{bold}{\encodingdefault}{\sfdefault}{bx}{n}

\def\gE{{\mathcal{E}}}

\def\gG{{\mathcal{G}}}

\def\gV{{\mathcal{V}}}

\def\sR{{\mathbb{R}}}

\DeclareMathOperator*{\argmin}{arg\,min}